\documentclass[10pt,twocolumn,letterpaper]{article}

\usepackage[pagenumbers]{cvpr} 

\usepackage[dvipsnames]{xcolor}

\usepackage{slashbox}

\usepackage[utf8]{inputenc} 
\usepackage[T1]{fontenc}    
\usepackage{url}            
\usepackage{booktabs}       
\usepackage{amsfonts}       
\usepackage{nicefrac}       
\usepackage{microtype}      
\usepackage{xcolor}         

\usepackage{graphicx}
\usepackage{amsmath}
\usepackage{amssymb}
\usepackage{booktabs}
\usepackage{smile}
\usepackage{xcolor}
\usepackage{bbm}
\usepackage{pifont}
\usepackage[export]{adjustbox}
\usepackage{wrapfig}
\usepackage[most]{tcolorbox}
\usepackage{amsmath, mathtools}

\newcommand\tf[1]{\textbf{#1}}

\def\ie{\textit{i.e.}}
\def\eg{\textit{e.g.}}

\def\etc{{\em etc.}}

\newcommand{\myparagraph}[1]{\vspace{1pt}\noindent{\bf{#1}}~~}
\makeatletter
\renewcommand{\paragraph}{%
  \@startsection{paragraph}{4}%
  {\z@}{0em}{-1em}%
  {\normalfont\normalsize\bfseries}%
}
\makeatother

\usepackage{colortbl}
\definecolor{lightgray}{gray}{0.75}
\definecolor{lightergray}{gray}{0.85}
\definecolor{Blue}{RGB}{3, 31, 97}
\definecolor{Blue1}{RGB}{214, 235, 245}
\definecolor{Blue2}{RGB}{235, 245, 250}
\definecolor{Gray}{RGB}{247, 252, 255}

\definecolor{convcolor}{HTML}{412F8A}
\definecolor{resnetcolor}{HTML}{8DA0CB}
\definecolor{vitcolor}{HTML}{fc8e62}

\newcommand{\convcolor}[1]{\textcolor{convcolor}{#1}}
\newcommand{\vitcolor}[1]{\textcolor{vitcolor}{#1}}

\definecolor{aliceblue}{rgb}{0.94, 0.97, 1.0}
\newcommand{\vb}{\vitcolor{$\mathbf{\circ}$\,}}
\newcommand{\cb}{\convcolor{$\bullet$\,}}
\newcommand{\gr}{\rowcolor[gray]{.95}}
\newcommand{\cgr}{\cellcolor[gray]{0.95}}

\usepackage{xspace}

\newcommand{\cfr}{\textrm{CFR}\xspace}

\newcommand{\waterbirds}{\textbf{Waterbirds}\xspace}
\newcommand{\celeba}{\textbf{CelebA}\xspace}

\newcommand{\metashift}{\textbf{MetaShift}\xspace}

\newcommand{\chexpert}{\textbf{CheXpert}\xspace}

\newcommand{\customfootnotetext}[2]{{
\renewcommand{\thefootnote}{#1}
\footnotetext[0]{#2}}}

\usepackage{tablefootnote}
\renewcommand\thefootnote{\textcolor{red}{\arabic{footnote}}}

\usepackage[toc,page,header]{appendix}
\usepackage{minitoc}

\usepackage[colorlinks,linkcolor=red,anchorcolor=blue,citecolor=teal,urlcolor=magenta]{hyperref}

\definecolor{cvprblue}{rgb}{0.21,0.49,0.74}

\def\paperID{1875} 
\def\confName{CVPR}
\def\confYear{2024}

\title{Calibrating Multi-modal Representations: \\ A Pursuit of Group Robustness without Annotations}

\author{Chenyu You\textsuperscript{$\dagger$},  Yifei Min\textsuperscript{$\dagger$}, Weicheng Dai\textsuperscript{$\dagger$}, \\ Jasjeet S. Sekhon, Lawrence Staib, 
James S. Duncan \\
Yale University \\
}

\begin{document}

\doparttoc 
\faketableofcontents 

\maketitle

\customfootnotetext{}{${\dagger}$ equal contribution. }

\begin{abstract}
Fine-tuning pre-trained vision-language models, like CLIP, has yielded success on diverse downstream tasks. However, several pain points persist for this paradigm: (i) directly tuning entire pre-trained models becomes both time-intensive and computationally costly. Additionally, these tuned models tend to become highly specialized, limiting their practicality for real-world deployment; (ii) recent studies indicate that pre-trained vision-language classifiers may overly depend on spurious features -- patterns that correlate with the target in training data, but are not related to the true labeling function; and (iii) existing studies on mitigating the reliance on spurious features, largely based on the assumption that we can identify such features, does not provide definitive assurance for real-world applications. As a piloting study, this work focuses on exploring mitigating the reliance on spurious features for CLIP without using any group annotation. 
To this end, we systematically study the existence of spurious correlation on CLIP and CLIP+ERM. 
We first, following recent work on Deep Feature Reweighting (DFR), verify that last-layer retraining can greatly improve group robustness on pretrained CLIP. 
In view of them, we advocate a lightweight representation calibration method for fine-tuning CLIP, by first generating a calibration set using the pretrained CLIP, and then calibrating representations of samples within this set through contrastive learning, all without the need for group labels. 
Extensive experiments and in-depth visualizations on several benchmarks validate the effectiveness of our proposals, largely reducing reliance and significantly boosting the model generalization. 
Our codes will be available in {{\href{https://github.com/charlesyou999648/cfr}{here}}}

\end{abstract}    
\section{Introduction}
\label{sec:intro}
In recent years, large-scale pre-trained vision-language models (VLMs) \cite{radford2021learning,jia2021scaling,li2021align,wang2021simvlm,zellers2021merlot,bain2021frozen} have showcased impressive capabilities across various downstream tasks, including visual understanding \cite{gu2021open,rao2022denseclip,lai2023scarcity,lai2023padclip}, image-text generation \cite{patashnik2021styleclip,mokady2021clipcap} and more \cite{guzhov2022audioclip,zhang2022pointclip,lai2023clipath}. Leveraging these well-learned and rich representations, fine-tuning pre-trained VLMs has been the dominant methodology, starting from pre-training from extensive web-crawled data and then tuning it towards specific downstream tasks. However, when relying on standard Empirical Risk Minimization (ERM) for training, it raises a risk of inadvertently amplifying spurious correlations, which may compromise robustness, especially for underrepresented groups \cite{geirhos2020shortcut}. 
Consequently, even these advanced VLMs are not exempt from the challenges posed by \textit{spurious correlation}, where patterns may correlate with the target class without truly pertaining to the classification function. 
This can inadvertently sideline certain minority groups within the training data, which poses practical challenges and limits the efficacy of these models in safety-critical applications. 
For example, on Waterbirds dataset \cite{sagawa2019distributionally}, tasked with classifying ``landbird'' and ``waterbird'', there exists a bias where an ``water (land)'' background is spuriously correlated with the ``waterbird (landbird)'' class, leading to a minority groups of ``waterbird on land'' and ``landbird on water''.

As a result, a considerable body of work has aimed at improving \textbf{group robustness} of \textit{vision} models \cite{sagawa2019distributionally,nam2022spread,liu2021just,creager2021environment,nam2020learning,izmailov2022feature,kirichenko2022last}. 
Specifically, to ensure learning models do not depend on spurious correlations, which can lead to high error rates in certain data groups, we align with mainstream practices and focus on enhancing the \textit{worst-case accuracy} (WGA) across different groups.
Yet, this remains underexplored in the \textit{vision-language} context. We therefore attempt to explore the central question motivating this work:
\vspace{-0.1em}
\begin{tcolorbox}[before skip=0.2cm, after skip=0.2cm, boxsep=0.0cm, middle=0.1cm, top=0.1cm, bottom=0.1cm]
\textit{\textbf{(Q)} Does there exist an efficient way to mitigate multi-modal spurious correlations without retraining the entire model, thereby enhancing its group robustness without relying on any group annotations?}
\end{tcolorbox}
\vspace{-0.1em}

To address \textbf{(Q)}, we aims to achieve two key objectives: 
({i}) \textbf{\textit{parameter-efficient fine-tuning}} -- traditional \textit{fine-tuning} typically involves updating a large proportion of or even all the parameters of the pre-trained model. 
Yet, in many practical scenarios, this paradigm becomes challenging owing to the significant memory and computational demands; and ({ii}) \textbf{\textit{group label efficiency}} -- since many real-world problems inherently contain spurious correlations, the existing methods often require prior group information to adapt large-scale pre-trained models for specific downstream datasets, which poses impediments to the deployment in real-world resource-constrained settings. 
Additionally, even when such spurious features are identifiable, the task of annotating vast datasets with group labels becomes prohibitively demanding.

In this paper, our research trajectory unfolds as follows: the \underline{\textbf{first}} part of our work provides comprehensive analysis to ascertain the presence of spurious correlations within CLIP:
\begin{itemize}
    \item \textbf{\textit{Identifying spurious correlation issues in CLIP:}} Our investigation uncovers spurious correlation issues within the large pre-trained multi-modal models.
    In plain words (detailed analysis in Sec.~\ref{sec:stage1}), we use \textit{t}-SNE \cite{van2008visualizing} and UMAP \cite{mcinnes2018umap} to inspect group-wise embeddings from benchmark datasets such as Waterbirds and CelebA. 
    Our findings clearly indicate the presence of unintended spurious correlations in both the pre-trained CLIP and CLIP+ERM representations. 
    This limitation arises from an over-reliance on spurious features, indicating a need for more robust feature calibrator tailored to specific downstream tasks.
    \item \textbf{\textit{Verifying the efficacy of feature extractor in CLIP:}} To tackle the challenge of spurious correlations without incurring substantial computational overhead, we draw inspiration from \cite{kirichenko2022last} to calibrate feature representation quality by re-training of CLIP's final layer.
    This approach allows seamless adaptation of large pre-trained multi-modal models to specific downstream benchmarks. 
    Empirical results suggest the feasibility of recalibrating the pre-trained feature representation to neutralize spurious correlations. 
    These findings further motivates us to explore more parameter-efficient methods devoid of group information reliance.
\end{itemize}

The \underline{\textbf{second}} part of our work is to put forth a streamlined approach to address the aforementioned concerns. 
To this end, we present a robust representation calibration approach that functions \textit{without the need for any group annotations}. 
This novel framework, \underline{\textbf{C}}ontrastive \underline{\textbf{F}}eature \underline{\textbf{R}}ecalibration (\textbf{\cfr}), integrates a contrastive learning paradigm into representation calibration, as shown in Figure \ref{fig:model} (Appendix).
\begin{itemize}
    \item Using the pre-trained CLIP, we construct a calibration set curated from the training data. The samples within this set act as pivotal anchor points. By calibrating the representation of these anchors, we aim to enhance robustness across the entire dataset. In curating this set, \cfr employs an intuitive strategy, selecting samples misclassified by the pre-trained CLIP.
    \item With the calibration set established, \cfr refines the sample representations, aligning them more closely with the centroid of their designated class in the feature space and distancing them from opposing class centroids. This calibration is adeptly achieved using a contrastive loss.
\end{itemize}

Our extensive experiments on CLIP (underexplored in semi-supervised spurious correlation literature till date) across multiple datasets illustrate that \cfr not only significantly improves group robustness compared to semi-supervised methods but also rivals the performance of supervised approaches. 
Furthermore, utilizing \textit{t}-SNE and UMAP, we observe that our proposed method exhibits substantially better class separation patterns compared to the pre-trained CLIP and CLIP fine-tuned with ERM.
In addition, referring to the training-validation curves of different methods across four benchmark datasets (Figure~\ref{fig:training validation curve}), it becomes evident that \cfr maintains its superiority in the ability to converge towards an optimal solution when compared to other methods.
Collectively, these experiments provide strong support for the efficacy of \cfr in addressing spurious correlations, all without reliance on group-specific information.

\begin{figure*}[t]
\centering
\includegraphics[width=0.92\linewidth]{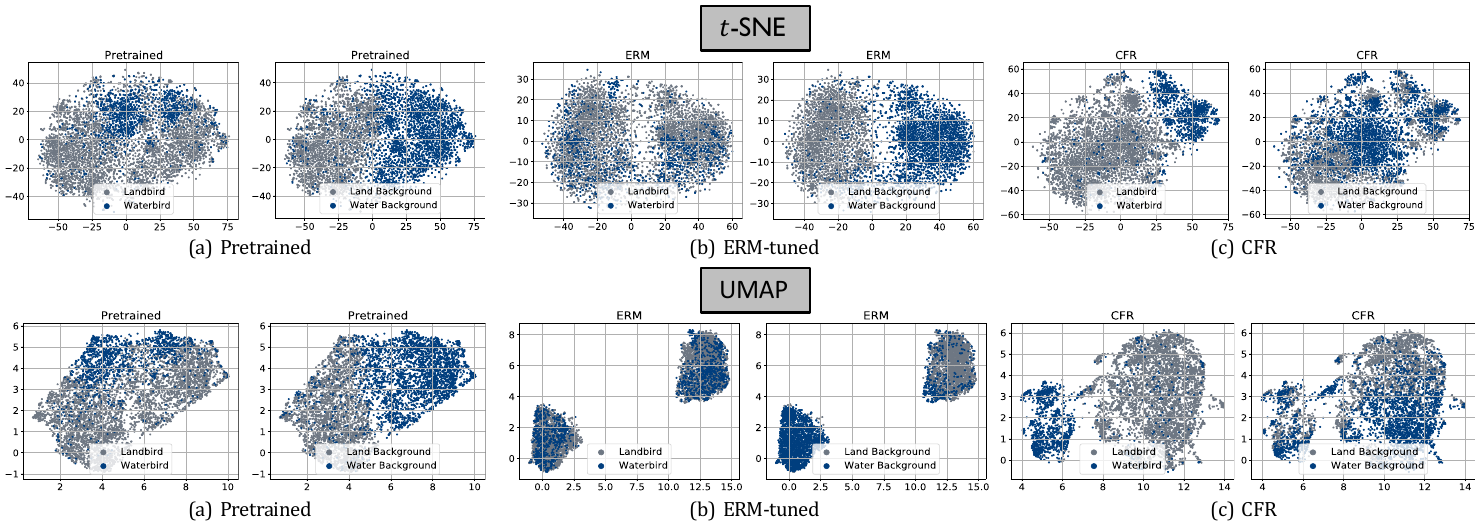}
\vspace{-5pt}
\caption{
\textbf{$t$-SNE and UMAP visualizations for pre-trained CLIP, ERM-tuned CLIP, and \cfr (ours) on Waterbirds.} 
We observe that both the pre-trained and ERM-tuned CLIP exhibit noticeable spurious correlations, with feature separations inappropriately aligned with spurious attributes, specifically the background, rather than the target class. In contrast, our method, as visualized through $t$-SNE and UMAP, demonstrates a significantly improved class separations, underscoring the robustness of our method in reducing spurious correlations.
} 
\label{fig:tsne umap waterbirds 3 methods}
\vspace{-10pt}
\end{figure*}
\section{Related Work}
\label{sec:related}
\paragraph{On spurious correlations.} 
There has been recently a burgeoning interest in examining the role of spurious correlations in the context of deep learning, particularly as it pertains to a wide range of real-world challenges. 
Existing work \cite{geirhos2020shortcut} shows that neural networks typically exhibit an inherent inclination to emphasize the intended features over the shortcut -- shallow features that are spuriously correlated with the classification targets, which may be particularly problematic in high-stakes and safety-critical scenarios. 
In vision, models often hinge on semantically irrelevant attributes such as an image's background \cite{sagawa2019distributionally,xiao2020noise,lai2023clipath,moayeri2022comprehensive}, texture \cite{geirhos2018imagenet}, and secondary objects \cite{rosenfeld2018elephant,shetty2019not,singla2021salient}, and other semantically irrelevant features \cite{li2018resound,brendel2019approximating,he2021interpretable}. 
Particularly concerning is their use in high-stakes areas like medical imaging, where networks might erroneously focus on hospital-specific tokens \cite{zech2018variable} or incidental cues \cite{oakden2020hidden} rather than actual disease symptoms. 
Similarly, in natural language processing (NLP), pre-trained models often exhibit a reliance on superficial features. 
This reliance allows them to perform well on benchmarks, even when not genuinely comprehending the tasks. 
For instance, models might leverage basic syntactic patterns, like lexical similarities between sentences, to deduce their interrelationship \cite{kaushik2018much,gururangan2018annotation,niven2019probing,mccoy2019right}. Broader introductions are provided in \cite{geirhos2020shortcut,yang2023change}. 
In this paper, we take a step further to meet more practical requirements -- \emph{without relying on any group annotations}, but with the help of language attributes on a pre-trained VLM with fine-tuning contrastive feature calibration. 
To our best knowledge, this training paradigm is underexplored in the spurious correlation literature,  offering a new view to improve group robustness of large-scale pre-trained multi-modal models.

\paragraph{Robustness and group annotations.}
Group robustness has recently garnered significant attention owing to the prevalence of spurious correlations, and hereby we focus our discussion on closely related works, both those that rely on group annotations and those that do not. 
({i}) \textit{Annotation-dependent methods:} A body of research has explored leveraging the group annotation. Much of this has aimed to improve robustness based on some heuristics like minimizing the worst-group loss \cite{sagawa2019distributionally}, learn invariant or diverse features \cite{arjovsky2019invariant,goel2020model,lin2022cascade,zhang2022rich,xu2022controlling}, class or group balancing or weighting \cite{cui2019class,menon2020overparameterisation,lai2023scarcity,idrissi2022simple,kirichenko2022last,izmailov2022feature}, or contrastive learning \cite{taghanaki2021robust,yang2023mitigating}. It usually leads to competitive performance but is hard to deploy in the real-world applications due to annotation costs.
({ii}) \textit{Annotation-free methods:} Another line of work, which discards the use of group annotations, can be roughly categorized into various groups. For instance, existing studies leverage auxiliary models to pseudo-label the minority group or spurious features \cite{sohoni2020no,nam2020learning,yaghoobzadeh2019increasing,creager2021environment,kim2021biaswap,asgari2022masktune,kim2022learning,sohoni2021barack,zhang2022correct,lai2023padclip,labonte2023towards}.
Others have emphasized upweighting samples misclassified by an early-stopped model \cite{liu2021just}, reweighting or subsampling classes \cite{idrissi2022simple,qiu2023simple}, or employing robust losses and regularizations~\cite{pezeshki2021gradient,yang2022understanding,zhang2022contrastive}. 
These methods seek to bolster group robustness, albeit often necessitating held-out group annotations, dual-phase training, or the deployment of additional auxiliary models.
To the best of our understanding, the integration of feature recalibration with the absence of group labels has been underexplored for the efficient fine-tuning of pre-trained CLIP models.

\paragraph{Spurious correlation in vision-language models.}
The advent of transformer-based architecture has led to the development of advanced large-scale pre-trained multi-modal models, aiming to enhance the effectiveness of these models. Some prior work has introduced language features aimed at making vision classifiers more robust, including attention maps \cite{petryk2022guiding}, changes to feature attributes \cite{zhang2023diagnosing}. 
Several pioneering efforts \cite{zhang2022contrastive,yang2023mitigating} have been made to obtain pre-trained multi-modal models that are robust to spurious correlations. 
For example, \cite{zhang2022contrastive} designs a new contrastive adapter, integrating it with transfer learning to enhance group robustness.
Nevertheless, this approach does not always guarantee improved performance, particularly for specialized downstream tasks. 
\cite{yang2023mitigating} for the first time introduces a fine-tuning approach specifically tailored to mitigate spurious correlations in pre-trained multi-modal frameworks. 
In contrast to this work, our goal is to devise a pragmatic training paradigm that functions without relying on any group annotations.

Our approach stands distinct from prior studies, anchored in \textbf{two key insights}: (i) \textit{the pivotal role of feature recalibration in bolstering robustness without the need for group annotations}, and (ii) \textit{the advantageous influence of language attributes on vision classifiers' group robustness}.
Our approach directly refines the sample representations using a contrastive loss with specially sampled positive and negative batches. 
Consequently, we achieve enhanced robustness against spurious correlations, optimizing group accuracy, efficiency, and practicality.

\section{Preliminaries}
\label{sec:prelim}
\paragraph{Setting.}
In this work, we consider classification tasks within the group robustness setting \cite{sagawa2019distributionally}, wherein the input is denoted as $\mathbf{x}\!\in\!\mathcal{X}$ and target classes as $\mathbf{y}\!\in\!\mathcal{Y}$. Specifically, we assume the data distribution consists of multiple \textit{groups} ${g} \in \mathcal{G}$. Typically, these groups are defined by a combination of the class label $\mathbf{y} \in \mathcal{Y}$ and a spurious attribute ${s} \in \mathcal{S}$. Consider the Waterbirds dataset \cite{sagawa2019distributionally}. Here, the classification task involves classifying $\mathbf{y} \in \{\text{landbird}, \text{waterbird}\}$, and the background depicted in the image serves as the spurious attributes $s \in \{\text{land}, \text{water}\}$. Consequently, the groups are formulated from the combinations of the class label and the spurious attribute, denoted as $\mathcal{G} = \mathcal{Y} \times \mathcal{S}$.

The attribute $s$ is considered spurious when it correlates with $\mathbf{y}$ but lacks a causal relationship. For example, within the Waterbirds dataset, approximately 95\% of data points labeled as $\mathbf{y} = \text{waterbird}$ possess the spurious attribute $s = \text{water}$. 
Consequently, models trained on this dataset may heavily depend on the background (water) to predict the class (waterbird), leading to reduced performance on the minority group $g=(\text{water}, \text{landbird})$. 

To safeguard against models relying on spurious correlations, we align with mainstream practices~\cite{sagawa2019distributionally,nam2022spread,liu2021just,creager2021environment,nam2020learning,izmailov2022feature,kirichenko2022last}, and utilize \textit{worst group accuracy} (WGA) as our evaluation metric, which denotes the minimum predictive accuracy of our model across all groups.

\paragraph{Access to spurious attributes.}
Many current approaches addressing robustness against spurious correlations presuppose the availability of spurious attributes $s$ within the training data \citep{sagawa2019distributionally,yang2023mitigating} or at least within a subset of data designated for model training \cite{kirichenko2022last,nam2022spread,sohoni2021barack}. In contrast, we delve into a more challenging scenario where the group information remains inaccessible for {fine-tuning}.


\paragraph{CLIP.} Contrastive Language-Image Pre-training (CLIP) \cite{radford2021learning} \!\footnote{\url{https://github.com/openai/CLIP}} learns from over 400M image-caption pairs collected from the web \!\footnote{This dataset is not public.} by maximizing the similarity between the image and text. Specifically, CLIP consists of ({i}) a visual encoder, ({ii}) a text encoder, and ({iii}) the dot product of their outputs serves as ``alignment score'' between the input image and text. Formally, given a batch of \( N \) images and their associated captions, each image representation $\mathbf{v}$ should align with its corresponding text representation $\mathbf{u}$. The likelihood of image \( i \) aligning with caption \( j \) is expressed as \( \exp(\beta\mathbf{v}_i^T\mathbf{u}_j)/\sum_{k=1}^N \exp(\beta\mathbf{v}_i^T\mathbf{u}_k) \), with \( \beta \) being a hyperparameter \!\footnote{$\mathbf{v}_i$ and $\mathbf{u}_j$ are normalized prior to the dot product calculation.}.

\section{Representation of Pretrained Models}
\label{sec:stage1}

In this section, we investigate the inherent spurious correlations present in the CLIP model. Utilizing widely-used feature visualization techniques, including $t$-SNE, UMAP \cite{mcinnes2018umap}, and GradCAM \cite{selvaraju2017grad}, our objective is to substantiate the presence of spurious correlations within the CLIP framework. We chart our investigation as below:

(i) \textit{\textbf{Revealing spurious correlations in pre-trained CLIP:}} Our initial step involves illustrating spurious correlations in pre-trained CLIP. We utilize dimensionality reduction techniques such as $t$-SNE and UMAP to unveil the separation between class features and spurious attributes in both ERM and pre-trained CLIP. Our $t$-SNE visualization, as shown in Figure~\ref{fig:tsne umap waterbirds 3 methods}, indicates that the pre-trained CLIP model inadequately separates between classes, yet efficiently identifies spurious attributes, thus corroborating the existence of spurious correlations.
Remarkably, a similar pattern emerges when examining ERM-tuned CLIP. The resemblance between pre-trained CLIP and ERM-tuned CLIP in their $t$-SNE visualizations underscores the severe issue of spurious correlation in the pre-trained models, as ERM training is well-known to be prone to such correlations. The UMAP visualization further confirms our findings with a similar discernment (See Figure~\ref{fig:tsne umap waterbirds 3 methods}).

\begin{figure}[t]
\centering
\includegraphics[width=0.90\linewidth]{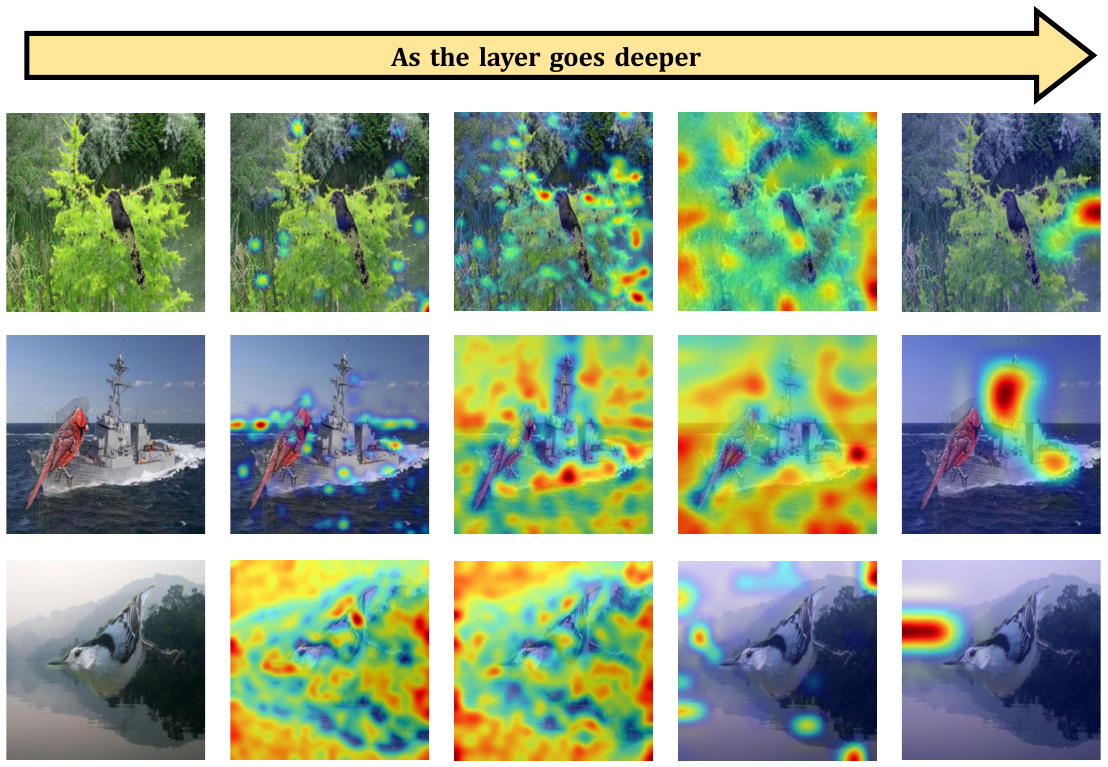}
\vspace{-5pt}
\caption{\textbf{Layer-by-layer GradCAM analysis of the CLIP-ResNet50.} Each row starts with the original image on the left, followed by four GradCAM visualizations corresponding to the four successive layers of the ResNet-50, with the depth of the layers increasing from left to right.}
\label{fig:gradcam evolve by layer}
\vspace{-10pt}
\end{figure}

(ii) \textit{\textbf{Investigating the underlying cause of spurious correlations:}} To further substantiate the presence of spurious correlation, we employ GradCAM to analyze various layers of the pre-trained CLIP model. This analysis reveals that pre-trained CLIP tends to focus on spurious attributes in the data, rather than the desired complex features (\eg, the bird) for classification.
In Figure~\ref{fig:gradcam evolve by layer}, we present the GradCAM results from each of the four layers of the ResNet-50 backbone in CLIP. A noticeable trend emerges: in shallower layers, the model's attention spans a broader region of the image. As we delve deeper into the layers, the model progressively narrows its focus to smaller regions of the image. Notably, in the final layer, the model concentrates its attention on an extremely limited portion of the image, which is often the background rather than the object of interest (\eg, the bird). This observation underscores the model's tendency to concentrate on spurious attributes, providing a plausible explanation for the prevalence of spurious correlation in pre-trained CLIP.

\textbf{Overall findings:} Our comprehensive examinations confirm the presence of spurious correlations in the features learned by the pre-trained CLIP. These insights prompt us to recalibrating CLIP model's features. Encouraged by these findings, we introduce our proposed approach, \textit{Calibrated Feature Refinement} ({\cfr}), in the subsequent section, aimed to mitigate the spurious correlation issue within the model.

\section{Feature Recalibration}
\label{sec:method}

In this section, we introduce \cfr, a novel representation calibration method, aimed at enhancing group robustness in pre-trained CLIP models, without the need for group annotations. As shown in Figure \ref{fig:model} (Appendix), \cfr unfolds in two pivotal steps: the initial step involves the assembly of a calibration set, thoughtfully curated from the training data. In the second step, we augment the robustness of the CLIP feature by calibrating the representation using the samples from the previously curated set.

\subsection{Calibration Set Formulation}
\label{sec:method calibration set formulation}

Sample selection strategies have been widely explored in the context of addressing spurious correlations~\cite{kirichenko2022last,liu2021just,zhang2022contrastive,zhang2022correct,labonte2023towards}. Such issues often arise when using the full training data. To mitigate this issue, researchers have focused on identifying more group-balanced subsets for refining models. 
In real-world scenarios where group information is commonly inaccessible, a widely accepted solution is to generate pseudo-group-labels, commonly achieved by using an auxiliary model trained with Empirical Risk Minimization (ERM). Existing works in this direction have taken three angles: (1) disagreement-based methods (\eg, selecting samples where ERM predictions contradict group-truth labels \cite{labonte2023towards}; (2) uncertainty-based method (\eg, opting for samples with high uncertainty in ERM predictions \cite{qiu2023simple}), and (3) clustering-based method (\eg, interpreting clusters formed by ERM-learned features as pseudo-groups \cite{sohoni2020no,zhang2022correct}), \etc

Inspired by these, we propose a straightforward yet highly effective approach tailored for vision-language models. Rather than training an entire model from the ground up using ERM, we opt to utilize the feature representation inherent in the pre-trained CLIP. Our strategy involves training only the projection head of CLIP using cross-entropy loss, subsequently selecting samples that are misclassified by this ERM-tuned CLIP for further rectification. 
Our empirical results underscore the efficacy of this approach, showcasing its impressive performance in recalibrating CLIP. Furthermore, we recognize the potential of exploring alternative selection methods as a promising avenue for future research.

\subsection{Contrastive Feature Recalibration}
\label{sec:method feature recalibration}

The subsequent phase is centered around recitifying the features of samples within this calibration set. By recalibrating these specific samples, our objective is to improve the quality of feature representation across the entire training dataset. Inspired by insights gained from the feature visualization of the pre-trained CLIP and ERM-tuned CLIP in Sec.~\ref{sec:stage1}, our proposed \cfr method is capable of recalibrating the feature of a given sample. This is achieved by pulling the sample's feature closer to the centroid of its designated class while simultaneously pushing it away from centroids associated with opposing classes.

Take an arbitrary sample $(\mathbf{x}, \mathbf{y})$ from the calibration set as the \textit{anchor}. We define $\mathbf{v}$ as the visual feature of $\mathbf{x}$, encoded by the pre-trained CLIP, situated prior to the final-layer projection head. Note that $\mathbf{v}$ remains constant throughout the recalibration process as updates are exclusively applied to the model parameters within the projection head. For ease of reference, we encapsulate a single sample with the tuple $(\mathbf{x}, \mathbf{y},\mathbf{v})$. 
Subsequently, $\mathbf{v}$ is processed through the projection head, denoted as $f_\theta(\cdot)$, with $\theta$ representing the parameter. This yields the final output feature $f_\theta(\mathbf{v})$. \cfr fine-tunes $f_{\theta}$, resulting in the recalibration of the output visual feature $f_\theta(\mathbf{v})$. It is crucial to note that $\mathbf{v}$ itself remains unaltered during training, as all layers preceding the projection head are frozen.
Let $c_\mathbf{y}$ represent the `optimal' centroid of class $\mathbf{y}$ within the feature space, which remains \textit{unknown} and is an idealized concept. 
The centroid of a class can be conceptualized as the geometric center of all samples pertaining to that class within the feature space. 
\cfr is designed to enhance the similarity between $f_\theta(\mathbf{v})$ and $c_\mathbf{y}$, while ensuring $f_\theta(\mathbf{v})$ remains distant from samples belonging to opposing classes. 

We would like to emphasize that these class centroids are unknown a priori, necessitating their estimation or on-the-fly learning during the training process. 
In the subsequent section, we present an in-depth discussion on the implementation of centroid estimation, employing various sample selection strategies. At a conceptual level, our approach draws inspiration from the state-of-the-art contrastive learning framework \cite{chen2020simple}, which guides our feature recalibration process. Specifically, we conduct feature recalibration by strategically selecting positive and negative samples within each batch, which will serve as the anchor points for the recalibration process. This method allows us to refine and enhance the features in a manner consistent with \cite{chen2020simple}.

\myparagraph{Estimation via Sample Selection.}
In our pursuit of refining the sample features within the calibration set, a na\"ive approach is to estimate the class centroid $c_\mathbf{y}$ as the mean representation of all samples within class $\mathbf{y}$, that is, $c_\mathbf{y} = \sum_{\mathbf{y}' = \mathbf{y}} f_\theta(\mathbf{v}') / M_\mathbf{y}$, where $M_\mathbf{y}$ represents the total sample count in class $\mathbf{y}' = \mathbf{y}$ within the training dataset.
However, this method has two significant limitations: (1) It does not ensure the accuracy of $c_\mathbf{y}$ as an optimal centroid, given that the averaging includes samples that may be incorrectly predicted, leading to potential misalignments in the feature space. (2) The iterative updates to $f_\theta(\cdot)$ render the exact computation resource-intensive.
To address the first concern, we refine our strategy to consider solely those samples in class $y$ accurately identified by the pre-trained model, ensuring $\hat{\mathbf{y}}' = \mathbf{y}$. 
From this curated subset, we randomly select a subset of data, denoted as $P(\mathbf{x})$ (with $P$ standing for ‘positive’), to mitigate the influence of misaligned samples.
To tackle the second (computational) challenge, we employ an Exponential Moving Average (EMA) update, formulating the class centroid updating scheme as follows:
\begin{align}
\label{eq: centroid EMA}
    c_\mathbf{y} \leftarrow (1-\gamma) c_\mathbf{y} + \gamma \sum_{\mathbf{x}' \in P(\mathbf{x})} f_{\theta} (\mathbf{v}')/|P(\mathbf{x})|,
\end{align}
where $|P(\mathbf{x})|$ denotes the cardinality of $P(\mathbf{x})$. 

To intensify the recalibration effect, we pair the positive subset $P(\mathbf{x})$ with the centroid $\{c_\mathbf{y}\}$ to form a positive mini-batch for a data point $(\mathbf{x}, \mathbf{y}, \mathbf{v})$.
That is, the positive mini-batch is given as $P(\mathbf{x}) \cup \{c_\mathbf{y}\}$.
We term this sample selection strategy for the positive mini-batch as \textit{\textbf{D}ynamic \textbf{P}ositive Centroid \textbf{S}ampling} (\textbf{DPS}). 

For distancing the feature representation $f_\theta(\mathbf{v})$ from different classes, \cfr selects a negative mini-batch $N(\mathbf{x})$ through two strategies: 
(1)
\textit{\textbf{R}andom \textbf{N}egative \textbf{S}ampling} (\textbf{RNS}): randomly choosing negative samples outside the anchor's class, and
(2) 
\textit{\textbf{N}earest-neighbour \textbf{N}egative \textbf{S}ampling} (\textbf{NNS}): selecting negative samples from the top-$k$ instances closest to the anchor within the feature space.\footnote{The feature space here refers to the feature before the final projection layer of CLIP's visual branch, represented as $\mathbf{v}$ (introduced in Sec.~\ref{sec:method feature recalibration}).}
While the latter strategy, initially validated by \cite{zhang2022contrastive}, is effective, it also incurs significant computational load on large datasets. To address this, we implement a batch sampling method followed by a top-$k$ selection within the batch.

Combining the positive and negative sample selection, we now have two options: \{DPS+RNS\} and \{DPS+NNS\}, which differ only in how the negative batch is selected.

\myparagraph{Calibration loss.} With the positive and negative batches in place, \cfr applies a contrastive loss for recalibration. 
For an individual instance $(\mathbf{x}, \mathbf{y}, \mathbf{v})$ as the anchor, its loss is:
\begin{align}
\label{eq:calibration loss}
    \mathcal{L}_{\textnormal{cal}} (\mathbf{x}) = - \frac{1}{|P(\mathbf{x})|+1}\quad \smashoperator{\sum_{\mathbf{v}^+\in P(\mathbf{x})\cup\{c_\mathbf{y}\}}} \ \log \frac{e^{z_+}}{e^{z_+} + \smashoperator{\sum_{\mathbf{v}^{-}\in N(\mathbf{x})}} e^{z_-}}, 
\end{align}
where each $z_{+}$ replies on $\mathbf{v}^{+}$ and is given by
$z_{+} \!=\! {\langle f_\theta(\mathbf{v}), f_\theta(\mathbf{v}^+) \rangle}/{\tau}$. 
Similarly $z_- = {\langle f_\theta(\mathbf{v}), f_\theta(\mathbf{v}^-) \rangle}/{\tau}$.
In other words, $\mathcal{L}_\textnormal{Cal}$ is a contrastive loss applied to the anchor $(\mathbf{x}, \mathbf{y}, \mathbf{v})$ with the positive and negative batch selected by either of the two strategies (\ie, DPS+RNS and DPS+NNS).

\myparagraph{Holistic Data Integration.}
Determining the size and distribution of the calibration set beforehand is a challenge, since the composition of this set can significantly differ across various downstream tasks.
Particularly when dealing with a limited-sized calibration set, there is a considerable risk of model overfitting if training relies solely on this small subset, jeopardizing the model’s ability to generalize.

Recognizing this potential pitfall, we incorporate the entire training dataset into the loss function for adjustment. 
Specifically, we start by selecting a mini-batch from the entire training dataset. For each sample $(\mathbf{x}, \mathbf{y}, \mathbf{v})$ within this mini-batch, we identify positive examples (those belonging to the same class) and negative examples (those from different classes).
To optimize the model, we adopt cosine similarity loss, aiming to reduce the distance between positive pairs while increasing the distance between negative pairs. The loss is formulated as follows:
\begin{align}\label{eq:spurious cosine similarity}
    & \mathcal{L}_\textnormal{CS} (\mathbf{x})= - \sum_{p=1}^P \frac{\mathbf{u}^\top \mathbf{u}_p}{\|\mathbf{u}\|\cdot\|\mathbf{u}_p\|} + \sum_{j=1}^J \frac{\mathbf{u}^\top \mathbf{u}_j}{\|\mathbf{u}\|\cdot\|\mathbf{u}_j\|},
\end{align} 
where $\mathbf{u}$, $\{\mathbf{u}_p\}_{p=1}^P$ and $\{\mathbf{u}_j\}_{j=1}^J$ denote the final visual representation of the image $\mathbf{x}$, images from the same group, and images from other groups, respectively. That is, $\mathbf{u}=f_\theta(\mathbf{v})$, $\mathbf{u}_p=f_\theta(\mathbf{v}_p)$, and $\mathbf{u}_j=f_\theta(\mathbf{v}_j)$.

\myparagraph{Final loss function.} The final loss combines the cosine similarity loss $\mathcal{L}_\textnormal{CS}$ (Eq.~\ref{eq:spurious cosine similarity}) and  the calibration loss $\mathcal{L}_\textnormal{cal}$ (Eq.~\ref{eq:calibration loss}):
\begin{align}\label{eq:final loss}
    \mathcal{L}_\textnormal{total} = \lambda \sum_{\mathbf{x}\in \mathcal{D}_\textnormal{cal}}  \mathcal{L}_{\textnormal{cal}} (\mathbf{x})   + \sum_{\mathbf{x}\in\mathcal{D}_\textnormal{all}}\mathcal{L}_\textnormal{CS} (\mathbf{x}),
\end{align}
where $\lambda$ is a tunable parameter and $\mathcal{D}_\textnormal{cal}$ and $\mathcal{D}_\textnormal{all}$ are the calibration set and the entire training set, respectively.
 \section{Experiments}
\label{sec:experiment}

\begin{table*}[t]
\begin{center}
    \caption{\textbf{Comparison results across various supervised methods, semi-supervised methods and our proposed four methods across the Waterbirds, CelebA, CheXpert and MetaShift benchmarks.} Best results within the \textit{semi-supervised} group are in {\tf{bold}}. Please refer to the text for discussion.
    }
	\vspace{-10pt}
\label{table:results_main}
\begin{adjustbox}{width=0.95\linewidth}
\begin{tabular}{cccccccccccccccccc}
	\toprule
	& & \multicolumn{8}{c}{\textbf{ResNet-50}} & \multicolumn{8}{c}{\textbf{ViT}}\\
	\cmidrule(r){3-10} \cmidrule(r){11-18}
	& & \multicolumn{2}{c}{Waterbirds} & \multicolumn{2}{c}{CelebA} & \multicolumn{2}{c}{CheXpert} & \multicolumn{2}{c}{MetaShift} & \multicolumn{2}{c}{Waterbirds} & \multicolumn{2}{c}{CelebA} & \multicolumn{2}{c}{CheXpert} & \multicolumn{2}{c}{MetaShift} \\ 
    \cmidrule(r){3-4} \cmidrule(r){5-6} \cmidrule(r){7-8} \cmidrule(r){9-10} \cmidrule(r){11-12} \cmidrule(r){13-14} \cmidrule(r){15-16} \cmidrule(r){17-18}
		&{Method}
		            & WGA
		            & Avg
		            & WGA
		            & Avg
		            & WGA
		            & Avg
		            & WGA
		            & Avg
		            & WGA
		            & Avg
		            & WGA
		            & Avg
                    & WGA
                    & Avg
                    & WGA
                    & Avg
		            \\ \midrule
        \multirow{5}{*}{\bf supervised}
        & {ERM~\cite{vapnik1991principles}}
                    & {45.64}
                    & {94.08}
                    & {52.78}
		            & {93.88}
                    & {18.46}
                    & {90.01}
                    & {73.85}
                    & {90.05}
                    & {57.91}
		            & {97.60}
                    & {23.33}
                    & {94.30}
                    & {14.07}
                    & {90.48}
                    & {89.23}
                    & {97.37}
                    \\
        &{GroupDRO~\cite{sagawa2019distributionally}}
                    & {75.08}
                    & {83.84}
                    & {84.09}
		            & {89.54}
                    & {68.29}
                    & {75.04}
                    & {83.19}
                    & {87.30}
                    & {90.82}
		            & {96.37}
                    & {88.33}
                    & {91.24}
                    & {67.02}
                    & {73.53}
                    & {93.85}
                    & {97.37}
                    \\
        &{S-CS~\cite{yang2023mitigating}}
                    & {77.51}
                    & {83.16}
                    & {75.24}
		            & {80.38}
                    & {67.34}
                    & {74.74}
                    & {81.15}
                    & {89.82}
                    & {89.09}
		            & {95.69}
                    & {86.11}
                    & {89.29}
                    & {65.26}
                    & {74.48}
                    & {92.31}
                    & {97.14}
                    \\
        &{S-CL~\cite{yang2023mitigating}}
                    & {75.23}
                    & {85.96}
                    & {75.56}
		            & {80.56}
                    & {64.49}
                    & {75.89}
                    & {81.54}
                    & {88.79}
                    & {89.93}
		            & {96.04}
                    & {87.78}
                    & {90.51}
                    & {66.26}
                    & {74.19}
                    & {93.14}
                    & {96.89}
                    \\
        &{DFR~\cite{kirichenko2022last}}
                    & {73.22}
                    & {83.82}
                    & {82.22}
		            & {91.57}
                    & {60.64}
                    & {74.96}
                    & {83.08}
                    & {88.33}
                    & {89.69}
		            & {97.80}
                    & {85.56}
                    & {90.80}
                    & {68.09}
                    & {76.59}
                    & {92.31}
                    & {97.03}
                    \\\midrule 
        \multirow{7}{*}{\bf semi-sup}
        &{AFR~\cite{qiu2023simple}}
                    & {48.38}
                    & {89.31}
                    & {53.44}
		            & {94.25}
                    & {45.21}
                    & {59.41}
                    & {76.92}
                    & {86.84}
                    & {73.42}
		            & {88.17}
                    & {70.00}
                    & {85.17}
                    & {48.72}
                    & {74.99}
                    & {90.31}
                    & {97.14}
                    \\
        &{JTT~\cite{liu2021just}}
                    & {61.68}
                    & {90.63}
                    & {60.16}
		            & {79.93}
                    & {45.89}
                    & {59.01}
                    & {78.46}
                    & {89.36}
                    & {83.64}
		            & {97.29}
                    & {75.56}
                    & {93.25}
                    & {50.95}
                    & {73.96}
                    & {91.21}
                    & {94.16}
                    \\
        &{CnC~\cite{zhang2022correct}}
                    & {61.21}
                    & {87.14}
                    & {63.89}
		            & {90.34}
                    & {45.10}
                    & {57.52}
                    & {78.31}
                    & {87.07}
                    & {84.49}
		            & {97.51}
                    & {79.22}
                    & {89.33}
                    & {58.89}
                    & {74.46}
                    & {92.15}
                    & {94.74}
                    \\
        &{Con-Adapter~\cite{zhang2022contrastive}}
                    & {69.89}
                    & {70.51}
                    & {63.98}
		            & {90.19}
                    & {42.78}
                    & {59.12}
                    & {77.92}
                    & {85.47}
                    & {86.14}
		            & {95.54}
                    & {76.11}
                    & {93.06}
                    & {49.59}
                    & {71.98}
                    & {91.29}
                    & {93.36}
                    \\
        & \cgr \vb DPS+RNS
                    & \cgr {\tf{76.93}}
                    & \cgr {77.61}
                    & \cgr {\tf{73.66}}
                    & \cgr {81.07}
                    & \cgr {\tf{54.44}}
                    & \cgr {62.76}
                    & \cgr {\tf{81.54}}
                    & \cgr {89.52}
                    & \cgr {\tf{88.23}}
		            & \cgr 96.79
                    & \cgr {\tf{84.77}}
                    & \cgr 87.81
                    & \cgr {\tf{64.11}}
                    & \cgr 73.48
                    & \cgr {\tf{93.72}}
                    & \cgr 95.54
                    \\ 
        &  \cgr \vb RPS+RNS
                    & \cgr 73.08
                    & \cgr 76.66
                    & \cgr 72.78
                    & \cgr 80.52
                    & \cgr 49.50
                    & \cgr 61.02
                    & \cgr 80.13
                    & \cgr 84.55
                    & \cgr 85.67
		            & \cgr 94.74
                    & \cgr 83.78
                    & \cgr 88.17
                    & \cgr 62.03
                    & \cgr 74.22
                    & \cgr 91.15
                    & \cgr 94.05
                    \\ 
        &  \cgr \cb DPS+NNS
                    & \cgr {{76.63}}
                    & \cgr 78.93
                    & \cgr {{73.21}}
                    & \cgr 77.52
                    & \cgr {{52.67}}
                    & \cgr 62.67
                    & \cgr {\tf{81.54}}
                    & \cgr 89.59
                    & \cgr {{87.58}}
		            & \cgr 96.40
                    & \cgr {{84.11}}
                    & \cgr 86.67
                    & \cgr {{63.69}}
                    & \cgr 71.62
                    & \cgr {{93.41}}
                    & \cgr 95.31
                    \\ 
        &  \cgr \cb RPS+NNS
                    & \cgr 72.43
                    & \cgr 77.26
                    & \cgr 68.44
                    & \cgr 70.99
                    & \cgr 46.25
                    & \cgr 60.96
                    & \cgr {81.15}
                    & \cgr 89.24
                    & \cgr 84.89
		            & \cgr 96.23
                    & \cgr 82.72
                    & \cgr 88.05
                    & \cgr 62.64
                    & \cgr 73.97
                    & \cgr 92.23
                    & \cgr 94.98
                    \\ 
		              \bottomrule
	\end{tabular}
    \end{adjustbox}
    \end{center}
    \vspace{-10pt}
\end{table*}

\begin{figure*}[t]
\centering
\includegraphics[width=0.86\linewidth]{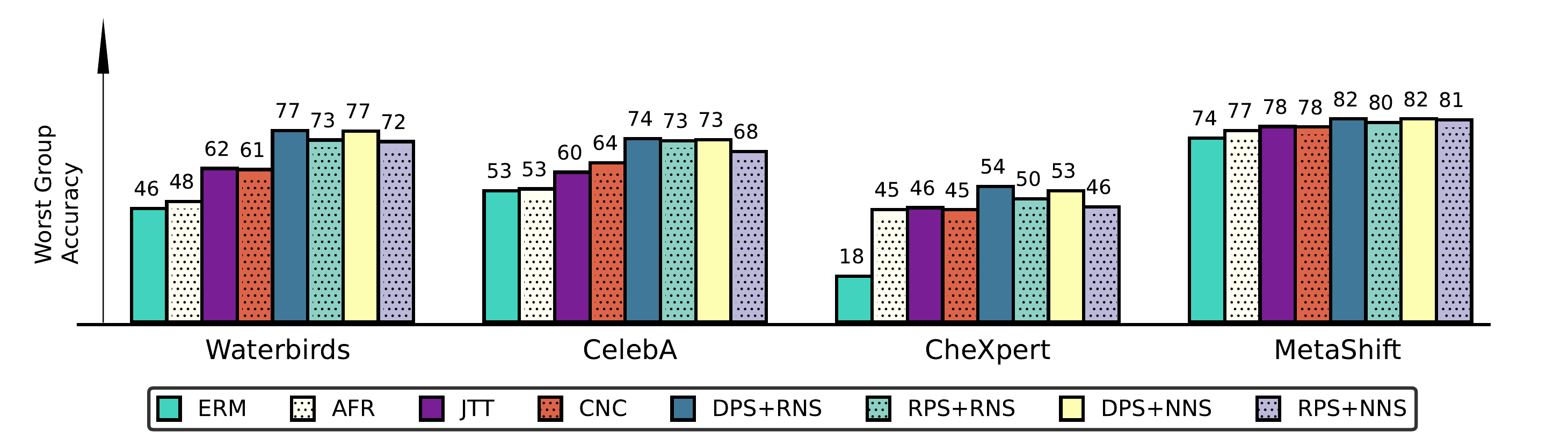}
\vspace{-5pt}
\caption{\textbf{Comparison of methods using the CLIP-ResNet50 architecture on four benchmark datasets.} We use Worst Group Accuracy to evaluate the performance for various methods, including ERM, semi-supervised baselines (\ie, AFR~\cite{qiu2023simple}, CnC~\cite{zhang2022correct}, JTT~\cite{liu2021just}), and our proposed methods.
We observe that \cfr combined with the sample selection strategies (\ie, \{DPS, RPS\}$\times$\{RNS,NNS\}) outperforms all semi-supervised baselines across all benchmarks.
} 
\label{fig:barplot semi without avg}
\vspace{-5pt}
\end{figure*}

In this section, we evaluate CFR across various benchmarks with spurious correlations, accompanied by comprehensive ablations on design choices and hyperparameter settings. 
Due to limited space, additional dataset and implementation details are discussed in Appendix~\ref{apdx:dataset} and \ref{apdx:implementation}.

\myparagraph{Evaluated methods.} 
As baseline methods, we compare against other semi-supervised methods, including JTT~\cite{liu2021just}, CnC~\cite{zhang2022correct} and AFR~\cite{qiu2023simple}. 
We also compare against methods that require group annotations, including GroupDRO~\cite{sagawa2019distributionally}, DFR~\cite{kirichenko2022last}, S-CS/S-CL~\cite{yang2023mitigating}. 

\myparagraph{Sampling Setup.} 
In Section~\ref{sec:method feature recalibration}, we detailed our DPS strategy for creating a positive batch for an anchor.
DPS combines $P(\mathbf{x})$, a subset of correctly predicted instances by pre-trained CLIP within the anchor's class $\mathbf{y}$, with $c_{\mathbf{y}}$, the Exponential Moving Average (EMA) estimated centroid of class $\mathbf{y}$ (Eq.~\ref{eq: centroid EMA}).
To assess the impact of incorporating the positive centroid $c_\mathbf{y}$ into the contrastive loss term $\mathcal{L}_{\textnormal{cal}}$, we propose to try \cfr with $c_\mathbf{y}$ removed from the loss term. 
Therefore, we explore an alternative to DPS, termed \textit{\textbf{R}andom \textbf{P}ositive \textbf{S}ampling} (\textbf{RPS}). 
RPS involves simply removing $c_\mathbf{y}$ from the positive batch of the contrastive loss $\mathcal{L}_{\textnormal{cal}}$, resulting in  a modified calibration loss term\footnote{This variant, compared to the DPS loss term (Eq.~\ref{eq:calibration loss}), simply excludes $\{c_\mathbf{y}\}$ from the summation.}: 
\begin{align*}
    \mathcal{L}_{\textnormal{cal}} (\mathbf{x}) &= - \frac{1}{|P(\mathbf{x})|} \quad \smashoperator{\sum_{\mathbf{v}^+\in P(\mathbf{x})}} \quad \log \frac{e^{z_+}}{e^{z_+} + \smashoperator{\sum_{\mathbf{v}^{-}\in N(\mathbf{x})}} e^{z_-}}. 
\end{align*}
Considering the setting of \{DPS, RPS\} \!$\times$\! \{RNS, NNS\}, we evaluate in total 4 sample selection strategies for \cfr: \{DPS+RNS\}, \{RPS+RNS\}, \{DPS+NNS\}, \{RPS+NNS\}.

\subsection{Results}
\label{sec:main results}

We now present the results of the methods across multiple benchmarks, using WGA as the performance metric. 

\begin{figure}[t]
\centering
\includegraphics[width=0.94\linewidth]{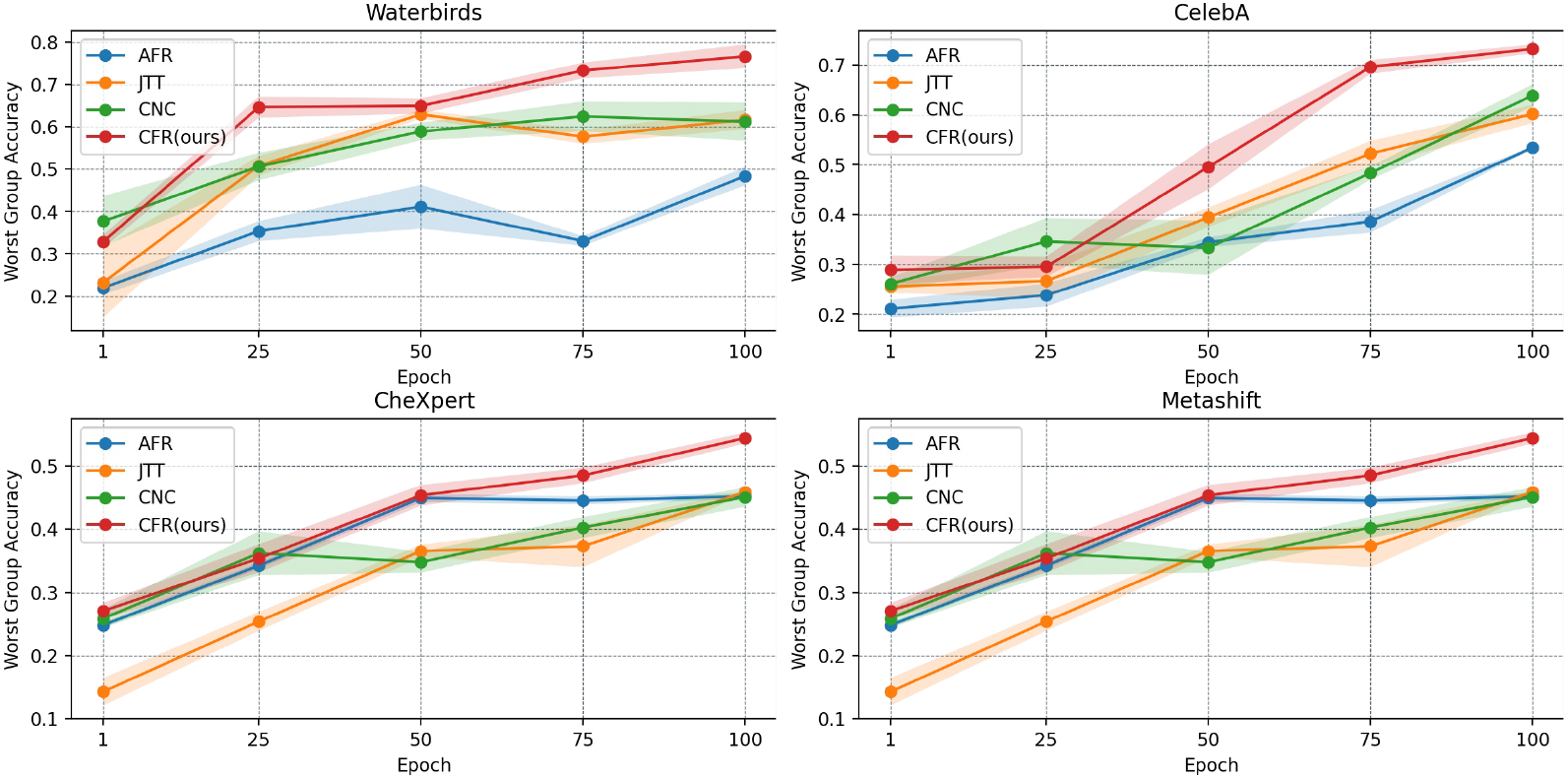}
\vspace{-5pt}
\caption{\textbf{Training-validation curves of various semi-supervised methods using CLIP-ResNet.} We plot WGA on a validation dataset at regular intervals of 25 epochs throughout the training process. 
Results are averaged across 3 random seeds. 
}
\label{fig:training validation curve}
\vspace{-5pt}
\end{figure}

\myparagraph{Main Results.}
We adopt classical visual backbones in CLIP, \ie, \!\{ResNet, ViT\}, and train
them in the combination of \textit{positive} and \textit{negative} sampling on \!\{DPS+RNS, DPS+NNS, RPS+RNS, RPS+NNS\}. Main results are in Table~\ref{table:results_main}, Figure~\ref{fig:barplot semi without avg}, and Figure \ref{fig:barplot vit semi without avg} (Appendix). Of note, we refer more discussion in Appendix \ref{apdx:additional results}
The following observations can be drawn: 
\ding{182}
Our \cfr demonstrates superior performance compared to all other semi-supervised training algorithms. 
Specifically, \cfr-ResNet with DPS+RNS obtains \{$15.25\!\sim\!28.55$, $9.77\!\sim\!20.22$, $8.55\!\sim\!9.34$, $3.08\!\sim\!4.62$\} WGA improvements across the datasets over AFR, JTT, CnC, respectively. 
This validates the effective of our proposed method. 
\ding{183} Appropriate selection strategies show consistent performance benefits across all two network backbones. Moreover, our DPS+RNS strategy surpasses the completely random strategy (RPS+RNS) in CLIP, which aligns well with our expectations as our assignment process is implicitly ``optimized'' by leveraging the naturally evolved feature embeddings.
\ding{184} When using ViT backbone, \cfr-ViT with DPS+RNS has up to \{$3.74\!\sim\!14.81$, $5.55\!\sim\!14.77$, $5.22\!\sim\!15.39$, $1.57\!\sim\!3.41$\} compared to all three semi-supervised baselines, respectively. 
\ding{185} As is shown in Table~\ref{table:results_main}, we observe that the improvements are more significant using ViT backbone than ResNet backbone. 
Our DPS+RNS approach the benchmarks set by fully supervised models. 
Besides, when using a ResNet backbone, a performance gap remains, suggesting room for further improvement. 
A possible reason is that, under the guidance of multi-modality information during fine-tuning, using ViT are less prone to capture spurious correlation features.
\ding{186}
By visualizing the training-validation curve\footnote{WGA results on the validation dataset during the training process.} (See Figure~\ref{fig:training validation curve}) of our method and other semi-supervised baselines, we observe that \cfr converges to a better optimal solution at a faster convergence rate, demonstrating the effectiveness of our approach.

\myparagraph{Analysis of Sample Selection.}
We conduct an extensive study to understand the performance benefits of different selection strategies in terms of WGA by comparing them against multiple state-of-the-art methods. Through comparison upon the following perspectives: (1) \textit{positive selection}: method w/o and w/ DPS strategy; (2) \textit{negative selection}: method w/o and w/ NNS strategy, we have the following findings.
\ding{182}
\textbf{DPS generally contributes to competitive performance gains.}
With the assistance of an adaptive sample selection process, DPS can lead to more robust and accurate models by effectively addressing challenges such as data imbalance and improving the model's focus on more minority-group features. 
As shown in Figure~\ref{fig:barplot semi without avg} and Table~\ref{table:results_main}, equipped with DPS+RNS, CLIP+ResNet achieves \{$3.85$, $0.88$, $4.94$, $1.44$\} WGA accuracy gains compared to RPS+RNS on Waterbirds, CelebA, CheXpert, and Metashift, respectively. 
Furthermore, we alter the model backbone to ViT, and observe a similar phenomenon happens to DPS+RNS with CLIP-ViT. 
A comparison between DPS+RNS and RPS+RNS shows that DPS+RNS performs better than RPS+RNS when using the ViT backbone.
Similarly, as in Figure~\ref{fig:barplot vit semi without avg} (Appendix) and Table~\ref{table:results_main}, we observe that DPS+NNS outperforms RPS+NNS in terms of WGA across four benchmarks.
This is clear evidence that DPS is able to better capture  minority-group than RPS.
\ding{183}
\textbf{In a principled way, RNS further boosts CLIP performance with DPS across all the evaluated benchmark datasets.}
Given the advancements of NNS, it is naturally expected that it would lead to better performance.
However, as depicted in Figure~\ref{fig:barplot semi without avg} and Table~\ref{table:results_main}, the utilization of DPS+RNS with CLIP-ResNet, compared to the baseline using DPS+NNS, achieves better performance across Waterbirds, CelebA, CheXpert, and Metashift.
Furthermore, our observations also reveal comparable improvements in WGA when utilizing CLIP-ViT (Appendix Figure~\ref{fig:barplot vit semi without avg}). This underscores the effectiveness of DPS+RNS compared to using DPS+NNS. 
Our findings suggest that employing RNS during the fine-tuning of CLIP-ResNet50 facilitates the model's ability to seeking the most informative features, particularly those are more correlated with minority-group features. We hypothesize that the observed decrease in effectiveness with NNS may be attributed to its sensitivity to the hyperparameter $k$. Unlike RNS, top-$k$ sampling is less random and, as indicated by \cite{zhang2022contrastive}, might require a larger batch size to be more effective. 
Nevertheless, it is important to note that a larger batch size for fine-tuning VLMs corresponds to increased computational resource requirements, which contradicts our objective of achieving lightweight fine-tuning.
We hope that our finding inspires future work to further explore utilizing larger batch sizes for NNS in VLMs when computational budget allows.

\subsection{Ablation}
\label{sec:ablation}
In this subsection, we conduct comprehensive ablation studies to gain deeper insights into the rationale behind our design choices. Note that all these experiments are conducted using the Waterbirds dataset with both CLIP-ResNet50 and CLIP-ViT models.

\myparagraph{Importance of Holistic Data Integration.}
We analyse the necessity of adding the loss component $\mathcal{L}_{\textnormal{CS}}$ from the Holistic Data Integration introduced in Sec.~\ref{sec:method feature recalibration}. 
We evaluate the setting  with and without $\mathcal{L}_{\textnormal{CS}}$ using WGA as the metric, as shown in Table~\ref{table:ablation no full data}. 
Our findings reveal a notable performance gains when employing Holistic Data Integration, particularly when using CLIP-ResNet50. This underscores the significant role of \(\mathcal{L}_{\textnormal{CS}}\) in achieving performance improvements, especially considering that the calibration loss term \(\mathcal{L}_{\textnormal{cal}}\) is confined to a relatively small calibration set.

\myparagraph{Extra Study.}
More studies on (1) weights of loss functions; (2) batch sizes in sample selection are in Appendix \ref{apdx:ablation}.

\begin{table}[t]
\caption{Ablation on the loss component $\mathcal{L}_{\textnormal{CS}}$ from Holistic Data Integration in Sec.~\ref{sec:method feature recalibration}. Adding $\mathcal{L}_{\textnormal{CS}}$ brings significant performance gain, especially with ResNet-50.} 
\label{table:ablation no full data}
\vspace{-5pt}
\centering
\resizebox{0.45\textwidth}{!}{
\begin{tabular}{@{\hskip 1mm}lcccccc@{\hskip 1mm}}
\toprule
& \multicolumn{3}{c}{\textbf{ResNet-50}} & \multicolumn{3}{c}{\textbf{ViT}}\\
\cmidrule(r){2-4} \cmidrule(r){5-7}
& \multicolumn{2}{c}{WGA} & & \multicolumn{2}{c}{WGA} &  \\ 
\cmidrule(r){2-3} \cmidrule(r){5-6}
Method  & {\cgr \vb w. $\mathcal{L}_\textnormal{CS}$ (\textbf{ours})} & {w/o $\mathcal{L}_\textnormal{CS}$} & {Gain$\uparrow$} & {\cgr \vb w. $\mathcal{L}_\textnormal{CS}$ (\textbf{ours})} & {w/o $\mathcal{L}_\textnormal{CS}$} & {Gain$\uparrow$} \\ 
\midrule
DPS+RNS  & \cgr 76.93 & 69.67 & 7.26 & \cgr 88.23 & 87.07 & 1.16 \\
RPS+RNS  & \cgr 73.08 & 65.98 & 7.10 & \cgr 85.67 & 85.23 & 0.44 \\
DPS+NNS  & \cgr 76.63 & 69.14 & 7.49 & \cgr 87.58 & 86.61 & 0.97 \\
RPS+NNS  & \cgr 72.43 & 70.40 & 2.03 & \cgr 84.89 & 83.02 & 1.87 \\
\bottomrule
\end{tabular}
}
\end{table}

\vspace{-5pt}

\section{Conclusion}
\label{sec:conclusion}
In our work, we study into the group robustness of the CLIP model without using any group annotations. Our initial findings indicate that retraining the last layer can considerably improve the group robustness of a pre-trained CLIP. Building upon this, we introduce a novel and efficient representation calibration technique for fine-tuning CLIP. This method involves creating a calibration set with the pre-trained CLIP and subsequently refining the representations of the samples within this set via contrastive learning, all without the need for group labels. Through comprehensive experiments and detailed visualizations across multiple benchmarks, our method demonstrates its capability to achieve state-of-the-art results in robust classification.
{
    \small
    \bibliographystyle{ieeenat_fullname}
    \bibliography{main}
}

\clearpage
\appendix

\clearpage
\setcounter{page}{1}

\maketitlesupplementary
\addcontentsline{toc}{section}{Appendix} 
\part{Appendix} 
{\hypersetup{linkcolor=black} \parttoc} 

\section{Datasets}
\label{apdx:dataset}

We describe the benchmarks details used in our study. We benchmark CFR on the following sources.
\begin{itemize}
    \item \waterbirds \cite{wah2011caltech}: is a widely-used binary classification dataset focusing on spurious correlations. This benchmark combines the Caltech-UCSD Birds-200-2011 (CUB) dataset \cite{wah2011caltech} with backgrounds from the Places dataset \cite{zhou2017places}. The classification goal discerns between landbirds and waterbirds, influenced by the spurious background attribute (either land or water). We adhere to the standard train/val/test splits as in \cite{idrissi2022simple}.
    \item \celeba \cite{liu2015deep} is a binary classification image dataset consisting of over 200,000 celebrity portraits. This dataset's pivotal task -- frequently cited in spurious correlation studies -- is to ascertain hair color (specifically distinguishing blond from non-blond). Intriguingly, gender emerges as the spurious attribute. We adhere to the standard dataset splits as in \cite{idrissi2022simple}. {This dataset is compliant with the \textit{Creative Commons Attribution 4.0 International} license.}.
    \item \chexpert \cite{irvin2019chexpert} is a comprehensive chest X-ray dataset from Stanford University Medical center, including over 200,000 images. The primary label is ``No Finding'', where a positive classification suggest the absence of illness. Drawing inspiration from \cite{seyyed2021underdiagnosis}, we factor in both race (\textit{White, Black, Other}) and gender as intertwined attributes. We adhere to the train/val/test splits as in \cite{yang2023change}.    
    \item \metashift \cite{liang2022metashift} stands as a versatile approach to crafting image datasets leveraging the Visual Genome project \cite{krishna2017visual}. In this work, we follow \cite{yang2023change} to adopt the pre-processed Cat \textit{vs.} Dog dataset, aiming to accurately identify these two distinct animal species. It is important to note the presence of a spurious attribute in this dataset -- image background, which tends to depict cats indoors and dogs outdoors. Further, we have utilized the ``unmixed'' version of the dataset, derived directly from the authors’ original codebase, ensuring reliability and integrity in our analysis.
\end{itemize}

\begin{figure*}[t]
\centering
\includegraphics[width=0.9\linewidth]{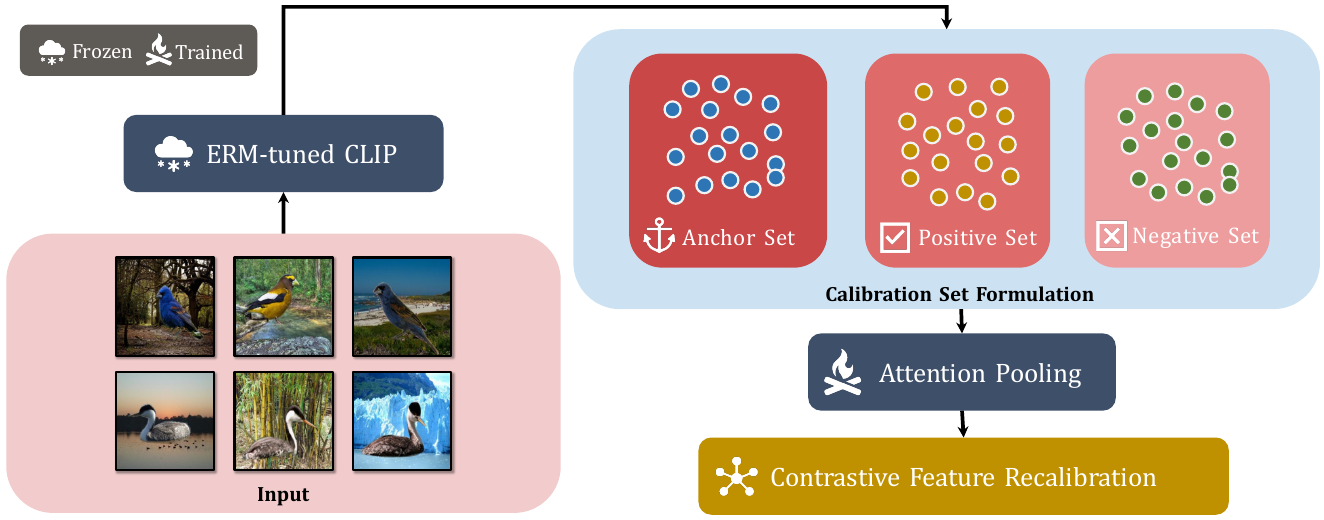}
\vspace{-5pt}
\caption{\textbf{The illustration of our proposed method \cfr.} \cfr decompose a  lightweight representation calibration into two steps. (1) \textbf{Calibration Set Formation}: When a pre-trained CLIP is fine-tuned using ERM, this ERM-tuned CLIP with the frozen weights constructs a calibration set from the training data, as introduced in Sec.~\ref{sec:method calibration set formulation} (Main Context). 
This set comprises pivotal anchor points, with each sample selected based on its misclassification by the ERM-tuned CLIP. These anchors play a crucial role in refining the robustness across the dataset. 
(2) \textbf{Contrastive Feature Recalibration}: Utilizing the curated calibration set, \cfr focuses on refining sample representations. 
This process involves aligning them more closely with the centroid of their respective class in the feature space while simultaneously distancing them from centroids of opposing classes. 
Such a recalibration is efficiently performed via a contrastive loss. Details about the positive and negative sample selection strategies used in \cfr are discussed in Sec.~\ref{sec:method feature recalibration} (Main Context).
} 
\label{fig:model}
\end{figure*}

\section{Implementation Details}
\label{apdx:implementation}

Here we give the full details of the implementation for all evaluated methods. A collections of hyper-parameters is given in Table~\ref{table:hyperparameter}.

\begin{table}[t]
\caption{\textbf{Experimental Settings}.} 
\label{table:hyperparameter}
\vspace{-5pt}
\centering
\resizebox{0.5\textwidth}{!}{
\begin{tabular}{@{\hskip 0.1mm}lccccc@{\hskip 0.1mm}}
\toprule
\textbf{Condition}  && \textbf{Parameter} && \textbf{Value} \\ 
\midrule
\textit{Model Architecture}: && && \\
\midrule
\multirow{3}{*}{CLIP-RN50~\cite{radford2021learning}} &&  Input size  &&  {256$\times$256} \\
  &&  Size of anchor/pos/neg features &&  $2048\times 7\times 7$ \\
  &&  Output size of projection layer &&   1024 \\
\midrule
\multirow{3}{*}{CLIP-ViT~\cite{radford2021learning}}  && Input size &&  $336\times 336$ \\
  && Size of anchor/pos/neg features &&  1024 \\
  && Output size of projection layer &&  768 \\
\midrule
\textit{Training}: && && \\
\midrule
\multirow{5}{*}{Optimizer}  &&  Type  &&  SGD \\
  &&  Learning rate &&  1e-5 \\
  &&  Momentum &&   0.9 \\
  &&  L2 weight decay &&   1e-4 \\
  && Metric to pick best model && WGA\\
\midrule
\multirow{2}{*}{Batch size}  && Anchor &&  128 \\
  && Cosine Similarity Loss &&  128 \\
\midrule
\midrule
\textit{Algorithm-specific}: && && \\
\midrule
\multirow{3}{*}{\vb \cfr (ours)}  &&  Number of positive points  &&  16 \\
  &&  Number of negative points &&  16 \\
  &&  EMA-coefficient (centroid) &&   0.9 \\
\midrule
\multirow{2}{*}{CnC~\cite{zhang2022correct}}  &&  Number of positive points &&  16 \\
  && Number of negative points &&  16 \\
\midrule
JTT~\cite{liu2021just}  &&  $\lambda_\text{up}$ &&  10 \\
\midrule
GroupDRO~\cite{sagawa2019distributionally}  &&  $\eta$ &&  0.01 \\
\midrule
\textit{Dataset-specific}: && && \\
\midrule
Waterbirds~\cite{wah2011caltech}  &&  Raw input size  &&  $224\times224$ \\
CelebA~\cite{liu2015deep}  &&  Raw input size  &&  $178\times218$ \\
CheXpert~\cite{irvin2019chexpert}  &&  Raw input size  &&  $390\times320$ \\
MetaShift~\cite{liang2022metashift}  &&  Raw input size  &&  $256\times256$ \\
\bottomrule
\end{tabular}
}
\end{table}

\paragraph{Model Architecture}
Our study utilizes CLIP~\cite{radford2021learning} as the visual-language model. CLIP comprises two different components: a visual and a language branch. Within the visual domain, we utilize two popular architectures, ResNets (RN) and Visual Transformers (ViT), with a specific focus on ResNet-50 (RN50) and ViT-L/14@336px, aligning with the setting in \cite{yang2023mitigating}. Here `ViT-L/14@336px' denotes the ViT-L/14 model that is fine-tuned on image inputs of 336 by 336 pixels. Concurrently, for the language branch, we incorporate the pre-trained mask language model, BERT~\cite{kenton2019bert}.

In adherence to established protocols from prior work \cite{yang2023mitigating}, our experiments, along with baseline methodologies, consistently freeze the language and vision encoders’s weights. This deliberate choice, aimed at advancing methods that are not only efficacious but also resource-efficient, allows for training solely on the projection layer. Such an approach preserves the intrinsic knowledge within the model and safeguards against model collapse (\ie, mitigating potential overfitting and yielding performance drops).

\myparagraph{Metrics.}
Our study utilizes `Worst-Group Accuracy' (WGA) and `Average Accuracy' (Avg) as key performance indicators. Specifically, WGA denotes the lowest model accuracy observed across diverse groups within the test dataset. These groups are determined by considering the product space of the spurious attributes and the classes within the test dataset. WGA is a widely adopted metric in the spurious correlation literature, providing insights into the model's robustness across different groupings. Meanwhile, Average Accuracy represents the classification accuracy averaged over all classes within the test set. It offers a holistic view of the model's overall performance across all class categories.

\myparagraph{Experimental Setup.}
In all of our experiments, we maintained consistent experimental setup using a single NVIDIA GeForce RTX 3090 GPU, and utilized fixed random seeds. We conduct our experiments using PyTorch 1.10.2+cu113 and Python 3.8.11, to ensure reproducibility.

\myparagraph{Calibration Set Generation.}
Here we describe how our methods generate the calibration set, an essential component of our research. The calibration set comprises tuples, each consisting of an anchor, its corresponding positive batch, and its negative batch. The process starts by selecting anchors from the training dataset. For each data point in the training set, we employ an ERM-tuned CLIP model to obtain an initial prediction. If the prediction is incorrect, the data point is added to the calibration set as an anchor.

To sample the positive batch for a given anchor, we pre-compute positive sets for each class. The positive set of a class comprises all training data that is correctly predicted by the ERM-tuned CLIP. This pre-computation of positive sets is performed only once at the beginning. To select the positive batch for an anchor, we sample a small positive batch of size $16$ uniformly from the positive set corresponding to the anchor's class.

For the negative batch, we also pre-compute negative sets, one for each class. For a specific class, the data points from all other classes are chosen to form the negative set. The key distinction is that the negative set includes data points regardless of the correctness of their initial predictions by the ERM-tuned CLIP. We have two options for building the negative batch from the negative set:
\begin{itemize}
    \item Option 1 ({RNS}): Randomly sample a small batch of size 16 from the negative set, matching the size of the positive batch. This option is referred to as the RNS option in the main text.
    \item Option 2 ({NNS}): Utilize cosine similarity to identify the nearest points among all the data points in the pre-computed negative set. Select the top 16 points to construct the negative batch, and this is referred to as the NNS option in the main text.
\end{itemize}

\myparagraph{Training Details.}
For all methods evaluated in our experiments, including both baselines and our approach, we employ an SGD optimizer with weight decay set to $10^{-4}$ and a momentum of $0.9$. The learning rate is fixed at $10^{-5}$, and the models are trained for 100 epochs. For the Calibration Loss ($\mathcal{L}_\textnormal{cal}$), the batch size is set to 128, implying that each batch consists of 128 anchors along with their corresponding positive and negative batches. For the Cosine Similarity Loss ($\mathcal{L}_{\textnormal{CS}}$), the batch size is also set to 128.

The model selection process remains consistent across all methods. We evaluate the model at the end of each epoch on the validation set and select the one with the best worst-group accuracy for the final testing. All accuracy metrics reported in this paper are based on the test set.

\myparagraph{Dataset Preprocessing.}
Our dataset preprocessing steps are same across all four datasets and all evaluated methods. Initially, we resize the raw images, maintaining a fixed height-to-width ratio. This ensures that the shorter edge of the image has dimensions of $256$ for ResNet-50 and $336$ for ViT-L/14@336px. Subsequently, the resized image is cropped to 256$\times$256 for ResNet-50 and 336$\times$336 for ViT-L/14@336px. Following this, the image is normalized by subtracting the average pixel value and dividing by the standard deviation, a procedure consistent with CLIP~\cite{radford2021learning}. No further data augmentation is applied after these steps, as our methods primarily focus on lightweight fine-tuning, which involves updating the projection layer of the vision branch of CLIP. Employing data augmentation in this scenario could lead to under-fitting due to the small parameter size of the projection layer.

\begin{figure*}[t]
\centering
\includegraphics[width=0.9\linewidth]{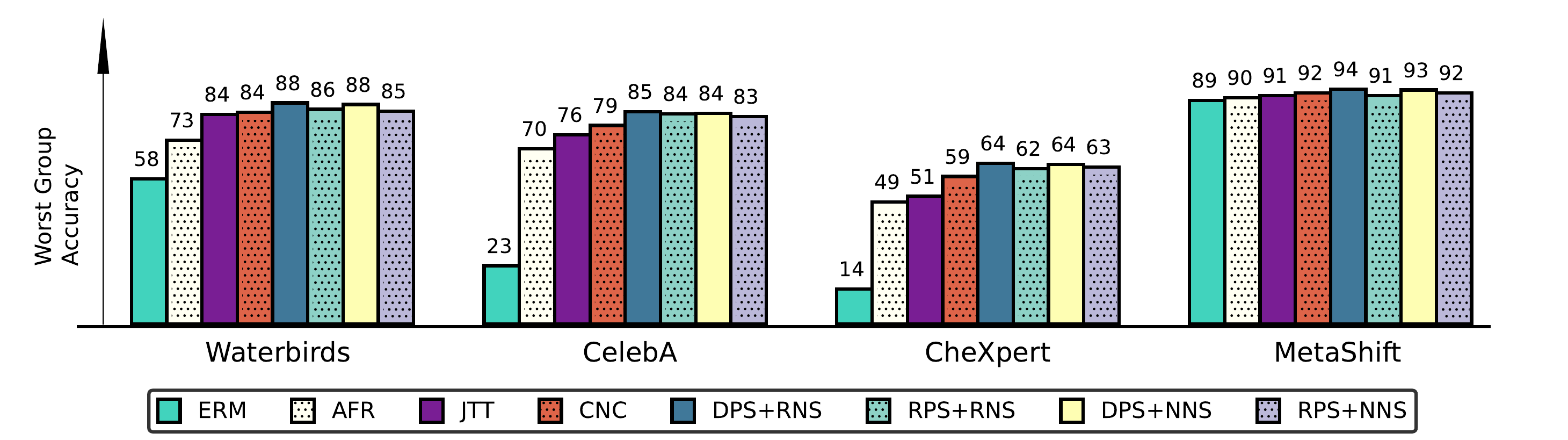}
\vspace{-5pt}
\caption{\textbf{Comparison of methods using the CLIP-ViT architecture on four benchmark datasets.} We use Worst Group Accuracy to evaluate the performance for various methods, including ERM, semi-supervised baselines (\ie, AFR~\cite{qiu2023simple}, CnC~\cite{zhang2022correct}, JTT~\cite{liu2021just}), and our proposed methods.
We observe that \cfr combined with the sample selection strategies (\ie, \{DPS, RPS\}$\times$\{RNS,NNS\}) outperforms all semi-supervised baselines across all benchmarks.
} 
\label{fig:barplot vit semi without avg}
\end{figure*}

\begin{table*}[t]
\begin{center}
    \caption{\textbf{Comparison results across various supervised methods, semi-supervised methods and our proposed four methods across the Waterbirds, CelebA, CheXpert, MetaShift and CMNIST (0.99) benchmarks.} Best results within the \textit{semi-supervised} group are in {\tf{bold}}. Please refer to the text for discussion.
    }
	\vspace{-10pt}
\label{table:results_main_with_cmnist}
\begin{adjustbox}{width=0.95\linewidth}
\begin{tabular}{cccccccccccccccccccccc}
	\toprule
	& & \multicolumn{10}{c}{\textbf{ResNet-50}} & \multicolumn{10}{c}{\textbf{ViT}}\\
	\cmidrule(r){3-12} \cmidrule(r){13-22}
	& & \multicolumn{2}{c}{Waterbirds} & \multicolumn{2}{c}{CelebA} & \multicolumn{2}{c}{CheXpert} & \multicolumn{2}{c}{MetaShift} & \multicolumn{2}{c}{CMNIST} & \multicolumn{2}{c}{Waterbirds} & \multicolumn{2}{c}{CelebA} & \multicolumn{2}{c}{CheXpert} & \multicolumn{2}{c}{MetaShift} & \multicolumn{2}{c}{CMNIST}\\ 
    \cmidrule(r){3-4} \cmidrule(r){5-6} \cmidrule(r){7-8} \cmidrule(r){9-10} \cmidrule(r){11-12} \cmidrule(r){13-14} \cmidrule(r){15-16} \cmidrule(r){17-18} \cmidrule(r){19-20} \cmidrule(r){21-22}
		&{Method}
		            & WGA
		            & Avg
		            & WGA
		            & Avg
		            & WGA
		            & Avg
		            & WGA
		            & Avg
		            & WGA
		            & Avg
		            & WGA
		            & Avg
                    & WGA
                    & Avg
                    & WGA
                    & Avg
                    & WGA
                    & Avg
                    & WGA
                    & Avg
		            \\ \midrule
        \multirow{6}{*}{\bf supervised}
        & {ERM~\cite{vapnik1991principles}}
                    & {45.64}
                    & {94.08}
                    & {52.78}
		            & {93.88}
                    & {18.46}
                    & {90.01}
                    & {73.85}
                    & {90.05}
                    & {0.00}
                    & {20.55}
                    & {57.91}
		            & {97.60}
                    & {23.33}
                    & {94.30}
                    & {14.07}
                    & {90.48}
                    & {89.23}
                    & {97.37}
                    & {52.43}
                    & {97.78}
                    \\
        & {Pretrained CLIP~\citep{radford2021learning}}
                    & {44.87}
                    & {90.81}
                    & {65.80}
		            & {80.74}
                    & {0.00}
                    & {80.44}
                    & {79.54}
                    & {91.74}
                    & {6.01}
                    & {52.06}
                    & {54.44}
		            & {93.46}
                    & {70.56}
                    & {85.04}
                    & {0.00}
                    & {89.96}
                    & {89.31}
                    & {97.14}
                    & {39.06}
                    & {71.31}
                    \\
        &{GroupDRO~\cite{sagawa2019distributionally}}
                    & {75.08}
                    & {83.84}
                    & {84.09}
		            & {89.54}
                    & {68.29}
                    & {75.04}
                    & {83.19}
                    & {87.30}
                    & {54.87}
                    & {69.19}
                    & {90.82}
		            & {96.37}
                    & {88.33}
                    & {91.24}
                    & {67.02}
                    & {73.53}
                    & {93.85}
                    & {97.37}
                    & {87.34}
                    & {94.17}
                    \\
        &{S-CS~\cite{yang2023mitigating}}
                    & {77.51}
                    & {83.16}
                    & {75.24}
		            & {80.38}
                    & {67.34}
                    & {74.74}
                    & {81.15}
                    & {89.82}
                    & {46.27}
                    & {59.62}
                    & {89.09}
		            & {95.69}
                    & {86.11}
                    & {89.29}
                    & {65.26}
                    & {74.48}
                    & {92.31}
                    & {97.14}
                    & {86.60}
                    & {95.54}
                    \\
        &{S-CL~\cite{yang2023mitigating}}
                    & {75.23}
                    & {85.96}
                    & {75.56}
		            & {80.56}
                    & {64.49}
                    & {75.89}
                    & {81.54}
                    & {88.79}
                    & {48.97}
                    & {58.75}
                    & {89.93}
		            & {96.04}
                    & {87.78}
                    & {90.51}
                    & {66.26}
                    & {74.19}
                    & {93.14}
                    & {96.89}
                    & {85.86}
                    & {95.29}
                    \\
        &{DFR~\cite{kirichenko2022last}}
                    & {73.22}
                    & {83.82}
                    & {82.22}
		            & {91.57}
                    & {60.64}
                    & {74.96}
                    & {83.08}
                    & {88.33}
                    & {48.97}
                    & {69.18}
                    & {89.69}
		            & {97.80}
                    & {85.56}
                    & {90.80}
                    & {68.09}
                    & {76.59}
                    & {92.31}
                    & {97.03}
                    & {83.87}
                    & {94.71}
                    \\\midrule 
        \multirow{7}{*}{\bf semi-sup}
        &{AFR~\cite{qiu2023simple}}
                    & {48.38}
                    & {89.31}
                    & {53.44}
		            & {94.25}
                    & {45.21}
                    & {59.41}
                    & {76.92}
                    & {86.84}
                    & {0.00}
                    & {21.39}
                    & {73.42}
		            & {88.17}
                    & {70.00}
                    & {85.17}
                    & {48.72}
                    & {74.99}
                    & {90.31}
                    & {97.14}
                    & {69.06}
                    & {71.34}
                    \\
        &{JTT~\cite{liu2021just}}
                    & {61.68}
                    & {90.63}
                    & {60.16}
		            & {79.93}
                    & {45.89}
                    & {59.01}
                    & {78.46}
                    & {89.36}
                    & {24.90}
                    & {41.81}
                    & {83.64}
		            & {97.29}
                    & {75.56}
                    & {93.25}
                    & {50.95}
                    & {73.96}
                    & {91.21}
                    & {94.16}
                    & {66.32}
                    & {82.27}
                    \\
        &{CnC~\cite{zhang2022correct}}
                    & {61.21}
                    & {87.14}
                    & {63.89}
		            & {90.34}
                    & {45.10}
                    & {57.52}
                    & {78.31}
                    & {87.07}
                    & {24.80}
                    & {66.50}
                    & {84.49}
		            & {97.51}
                    & {79.22}
                    & {89.33}
                    & {58.89}
                    & {74.46}
                    & {92.15}
                    & {94.74}
                    & {61.89}
                    & {82.48}
                    \\
        &{Con-Adapter~\cite{zhang2022contrastive}}
                    & {69.89}
                    & {70.51}
                    & {63.98}
		            & {90.19}
                    & {42.78}
                    & {59.12}
                    & {77.92}
                    & {85.47}
                    & {30.77}
                    & {41.89}
                    & {86.14}
		            & {95.54}
                    & {76.11}
                    & {93.06}
                    & {49.59}
                    & {71.98}
                    & {91.29}
                    & {93.36}
                    & {69.59}
                    & {84.48}
                    \\
        & \cgr \vb DPS+RNS
                    & \cgr {\tf{76.93}}
                    & \cgr {77.61}
                    & \cgr {\tf{73.66}}
                    & \cgr {81.07}
                    & \cgr {\tf{54.44}}
                    & \cgr {62.76}
                    & \cgr {\tf{81.54}}
                    & \cgr {89.52}
                    & \cgr {\tf{45.86}}
                    & \cgr {71.56}
                    & \cgr {\tf{88.23}}
		            & \cgr 96.79
                    & \cgr {\tf{84.77}}
                    & \cgr 87.81
                    & \cgr {\tf{64.11}}
                    & \cgr 73.48
                    & \cgr {\tf{93.72}}
                    & \cgr 95.54
                    & \cgr {\tf{76.92}}
                    & \cgr 90.44
                    \\ 
        &  \cgr \vb RPS+RNS
                    & \cgr 73.08
                    & \cgr 76.66
                    & \cgr 72.78
                    & \cgr 80.52
                    & \cgr 49.50
                    & \cgr 61.02
                    & \cgr 80.13
                    & \cgr 84.55
                    & \cgr{29.03}
                    & \cgr{63.61}
                    & \cgr 85.67
		            & \cgr 94.74
                    & \cgr 83.78
                    & \cgr 88.17
                    & \cgr 62.03
                    & \cgr 74.22
                    & \cgr 91.15
                    & \cgr 94.05
                    & \cgr 69.60
                    & \cgr 85.73
                    \\ 
        &  \cgr \cb DPS+NNS
                    & \cgr {{76.63}}
                    & \cgr 78.93
                    & \cgr {{73.21}}
                    & \cgr 77.52
                    & \cgr {{52.67}}
                    & \cgr 62.67
                    & \cgr {\tf{81.54}}
                    & \cgr 89.59
                    & \cgr{35.73}
                    & \cgr{69.60}
                    & \cgr {{87.58}}
		            & \cgr 96.40
                    & \cgr {{84.11}}
                    & \cgr 86.67
                    & \cgr {{63.69}}
                    & \cgr 71.62
                    & \cgr {{93.41}}
                    & \cgr 95.31
                    & \cgr 71.64
                    & \cgr 79.63
                    \\ 
        &  \cgr \cb RPS+NNS
                    & \cgr 72.43
                    & \cgr 77.26
                    & \cgr 68.44
                    & \cgr 70.99
                    & \cgr 46.25
                    & \cgr 60.96
                    & \cgr {81.15}
                    & \cgr 89.24
                    & \cgr{38.19}
                    & \cgr{65.54}
                    & \cgr 84.89
		            & \cgr 96.23
                    & \cgr 82.72
                    & \cgr 88.05
                    & \cgr 62.64
                    & \cgr 73.97
                    & \cgr 92.23
                    & \cgr 94.98
                    & \cgr 72.68
                    & \cgr 85.15
                    \\ 
		              \bottomrule
	\end{tabular}
    \end{adjustbox}
    \end{center}
    \vspace{-10pt}
\end{table*}

\section{Additional Results}
\label{apdx:additional results}
Owing to space limit in the main text, we have included Figure~\ref{fig:barplot vit semi without avg} in our supplementary materials. This figure provides a performance comparison of various semi-supervised methods utilizing {CLIP-ViT}, complementing Figure~\ref{fig:barplot semi without avg} from the main text, which specifically presents results for {CLIP-ResNet50}.

We report additional experiments on the Colored-MNIST (CMNIST) dataset for our methods and all the baselines. For CMNIST, we follow the same setup as in \citep{zhang2022correct}.
The result is shown in Table~\ref{table:results_main_with_cmnist}.
We also report the result on pretrained CLIP without any further training or fine-tuning. 
We observe that DPS+RNS still performs the best.

\begin{table}[t]
\caption{\textbf{Ablation on Different Loss Components.} In this analysis, we adopt the \{DPS+RNS\} sampling strategy.} 
\label{table:ablation loss ratio}
\vspace{-5pt}
\centering
\resizebox{0.5\textwidth}{!}{
\begin{tabular}{@{\hskip 1mm}lcccccc@{\hskip 1mm}}
\toprule
& \multicolumn{6}{c}{WGA} \\ 
\cmidrule(r){2-7}
Model  & 0.1 & 0.2 & 0.5 & \cgr \vb 1.0 (\textbf{ours}) & 2.0 & 5.0 \\ 
\midrule
CLIP-ResNet50  & 75.23 & 75.30 & 74.28 & \cgr 76.93 & 73.83 & 74.72 \\
CLIP-ViT  & 85.83 & 86.78 & 86.87 & \cgr 88.23 & 86.85 & 87.07 \\
\bottomrule
\end{tabular}
}
\end{table}

\begin{table}[t]
\caption{\textbf{Ablation on Batch Size in Sample Selection.} In this analysis, we adopt the \{DPS+RNS\} sample strategy.} 
\label{table:ablation batch size}
\vspace{-5pt}
\centering
\resizebox{0.5\textwidth}{!}{
\begin{tabular}{@{\hskip 1mm}lcccc|ccc@{\hskip 1mm}}
\toprule
& & \multicolumn{3}{c}{CLIP-ResNet50} & \multicolumn{3}{c}{CLIP-ViT} \\ 
\cmidrule(r){3-5} \cmidrule(r){6-8}
& & \multicolumn{3}{c}{Negative} & \multicolumn{3}{c}{Negative} \\
\cmidrule(r){3-5} \cmidrule(r){6-8}
& Size & 8 & 16 & 32 & 8 & 16 & 32 \\ 
\midrule
\multirow{3}{*}{Positive}& 8  & 73.79 & 73.83 & 73.05 & 86.76 & 86.89 & 87.07 \\
& 16  & 73.39 & 76.93 & 73.68 & 86.21 & 88.23 & 86.30 \\
& 32  & 72.99 & 72.43 & 73.36 & 86.60 & 86.43 & 86.45 \\
\bottomrule
\end{tabular}
}
\end{table}

\begin{table}[t]
\caption{\textbf{Ablation on the positive subset $P(\mathbf{x})$ on Waterbirds.} 
The configuration labeled as `$c_\mathbf{y}$-only+RNS' represents a scenario where $P(\mathbf{x})$ is excluded from the calibration loss, and RNS is employed as the method for selecting negative samples. For a more comprehensive comparison, we have also incorporated three semi-supervised baseline methods (\ie~ AFR, JTT, CnC). Notably, integrating $P(\mathbf{x})$ into the calibration loss results in a significant performance improvement for both ResNet-50 and ViT models.}
\label{table:positive subset}
\centering
\resizebox{0.38\textwidth}{!}{
\begin{tabular}{@{\hskip 1mm}lccccc@{\hskip 1mm}}
\toprule
& \multicolumn{2}{c}{\textbf{ResNet-50}} & \multicolumn{2}{c}{\textbf{ViT}}\\
\cmidrule(r){2-3} \cmidrule(r){4-5}
Method  & WGA & Avg & WGA & Avg  \\ 
\midrule
$c_\mathbf{y}$-only+RNS & 60.22 & 66.11 & 76.60 & 85.00 \\
\gr \vb DPS+RNS (ours) & 76.93 & 77.61 & 88.23 & 96.79
\\
\midrule
JTT~\cite{liu2021just} & 61.68 & 90.63 & 83.64 & 97.29
\\
CnC~\cite{zhang2022correct} & 61.21 & 87.14 & 84.49 & 97.51
\\
AFR~\cite{qiu2023simple} & 48.38 & 89.31 & 73.42 & 88.17
\\
\bottomrule
\end{tabular}
}
\end{table}


\begin{table}[t]
\caption{\textbf{Group information of Waterbirds.} 
The data pertains to the distribution of samples across all groups in the training and testing splits of the Waterbirds dataset. The dataset is categorized based on spurious attributes, which include \{Water Background (BG), Land BG\}, and the classes, which are \{Waterbird, Landbird\}.
}
\label{table:waterbirds group}
\centering
\resizebox{0.49\textwidth}{!}{
\begin{tabular}{@{\hskip 1mm}lcccccc@{\hskip 1mm}}
\toprule
&& \multicolumn{2}{c}{\textbf{Train}} & \multicolumn{2}{c}{\textbf{Test}}\\
\cmidrule(r){3-4} \cmidrule(r){5-6}
  && Water BG & Land BG & Water BG & Land BG  \\ 
\midrule
Waterbird && 1057 & 56 & 642 & 642 \\
Landbird && 184 & 3498 & 2255 & 2255
\\
\bottomrule
\end{tabular}
}
\end{table}


\begin{table}[t]
\caption{\textbf{Group information of CelebA.} 
The data pertains to the distribution of samples across all groups in the training and testing splits of the CelebA dataset. The dataset is categorized based on spurious attributes, which include \{Female, Male\}, and the classes, which are \{Blond, Non-blond\}.
}
\label{table:celebA group}
\centering
\resizebox{0.39\textwidth}{!}{
\begin{tabular}{@{\hskip 1mm}lcccccc@{\hskip 1mm}}
\toprule
&& \multicolumn{2}{c}{\textbf{Train}} & \multicolumn{2}{c}{\textbf{Test}}\\
\cmidrule(r){3-4} \cmidrule(r){5-6}
  && Female & Male & Female & Male  \\ 
\midrule
Blond && 22880 & 1387 & 2480 & 180 \\
Non-blond && 71629 & 66874 & 9767 & 7535
\\
\bottomrule
\end{tabular}
}
\end{table}


\begin{table*}[t]
\caption{\textbf{Group information of CheXpert.} 
The data pertains to the distribution of samples across all groups in the training and testing splits of the CheXpert dataset. The dataset is categorized based on spurious attributes, which include \{Black female, Black male, White female, White male, Other female, Other male\}, and the classes, which are \{Illness, No illness\}.
}
\label{table:chexpert group}
\centering
\resizebox{\textwidth}{!}{
\begin{tabular}{@{\hskip 1mm}lccccccccccccc@{\hskip 1mm}}
\toprule
&& \multicolumn{6}{c}{\textbf{Train}} & \multicolumn{6}{c}{\textbf{Test}}\\
\cmidrule(r){3-8} \cmidrule(r){9-14}
  && Black female & Black male & White female & White male & Other female & Other male & Black female & Black male & White female & White male & Other female & Other male  \\ 
\midrule
Illness && 3877 & 4051 & 33676 & 51606 & 23407 & 33604 & 783 & 796 & 6772 & 10420 & 4696 & 6763 \\
No illness && 506 & 543 & 3490 & 5446 & 2975 & 3912 & 94 & 123 & 661 & 990 & 581 & 740 
\\
\bottomrule
\end{tabular}
}
\end{table*}


\begin{table}[t]
\caption{\textbf{Group information of MetaShift.} 
This data pertains to the distribution of samples across all groups in the training and testing splits of the MetaShift dataset. The dataset is categorized based on spurious attributes, which include \{Indoor Background (BG), Outdoor BG\}, and the classes, which are \{Cat, Dog\}.
}
\label{table:metashift group}
\centering
\resizebox{0.42\textwidth}{!}{
\begin{tabular}{@{\hskip 1mm}lcccccc@{\hskip 1mm}}
\toprule
&& \multicolumn{2}{c}{\textbf{Train}} & \multicolumn{2}{c}{\textbf{Test}}\\
\cmidrule(r){3-4} \cmidrule(r){5-6}
  && Indoor BG & Outdoor BG& Indoor BG& Outdoor BG \\ 
\midrule
Cat && 630 & 153 & 345 & 65 \\
Dog && 402 & 635 & 191 & 273
\\
\bottomrule
\end{tabular}
}
\end{table}

\section{Additional Ablations}
\label{apdx:ablation}
\myparagraph{Ablation on Loss Function Weights.}
Our study delves into the balance between the two loss terms, \(\mathcal{L}_{\textnormal{cal}}\) and \(\mathcal{L}_{\textnormal{CS}}\), with a particular focus on the ratio \(\lambda\) as defined in Eq.~\eqref{eq:final loss}. The findings, as shown in Table~\ref{table:ablation loss ratio}, indicate that a \(\lambda\) value of 1.0 is optimal.

\myparagraph{Ablation on Batch Sizes for Sample Selection.}
As detailed in Sec.~\ref{sec:method feature recalibration}, our sample selection process involves selecting both a positive and a negative batch for each anchor within the calibration set. In this study, we delve into exploring the optimal batch sizes for these positive and negative samples. The result is shown in Table~\ref{table:ablation batch size}. We observe that our choice of $(16, 16)$ is the optimal for both the ResNet50 and ViT architectures.

\myparagraph{Ablation on Centroid-only Positive Batch.}
For the DPS sample selection strategy of the positive batch in Sec.~\ref{sec:method feature recalibration}, we incorporate the positive subset $P(\mathbf{x})$ into the calibration loss together with the estimated optimal centroid $c_\mathbf{y}$ to intensify the recalibration effect, as detailed by Eq.~\ref{eq:calibration loss}. 
To demonstrate the necessity of adding $P(\mathbf{x})$, we evaluate \cfr with $P(\mathbf{x})$ removed from the calibration loss.
This results in the following variant of the calibration loss, which simply removes $P(\mathbf{x})$ from the summation in Eq.~\ref{eq:calibration loss}:
\begin{align*}
    \mathcal{L}_{\textnormal{cal}} (\mathbf{x}) = -   \ \log \frac{e^{z_+}}{e^{z_+} + \smashoperator{\sum_{\mathbf{v}^{-}\in N(\mathbf{x})}} e^{z_-}}, 
\end{align*}
where $z_{+} \!=\! {\langle f_\theta(\mathbf{v}), c_\mathbf{y} \rangle}/{\tau}$, and $z_- = {\langle f_\theta(\mathbf{v}), f_\theta(\mathbf{v}^-) \rangle}/{\tau}$.
The ablation result is shown in Table~\ref{table:positive subset}.
Compared to semi-supervised baselines (\ie, AFR, JTT, CnC), our proposed feature recalibration method, utilizing only $c_\mathbf{y}$, outperforms or performs on par with the three semi-supervised baselines.
However, using $P(\mathbf{x})$ in the calibration loss (\ie~our default choice \{DPS+RNS\}) significantly boost the performance compared to $c_\mathbf{y}$-only, on both ResNet-50 and ViT.

\section{Additional Dataset Details}
\label{apdx:dataset detail}

\begin{figure}[t]
\centering
\includegraphics[width=0.97\linewidth]{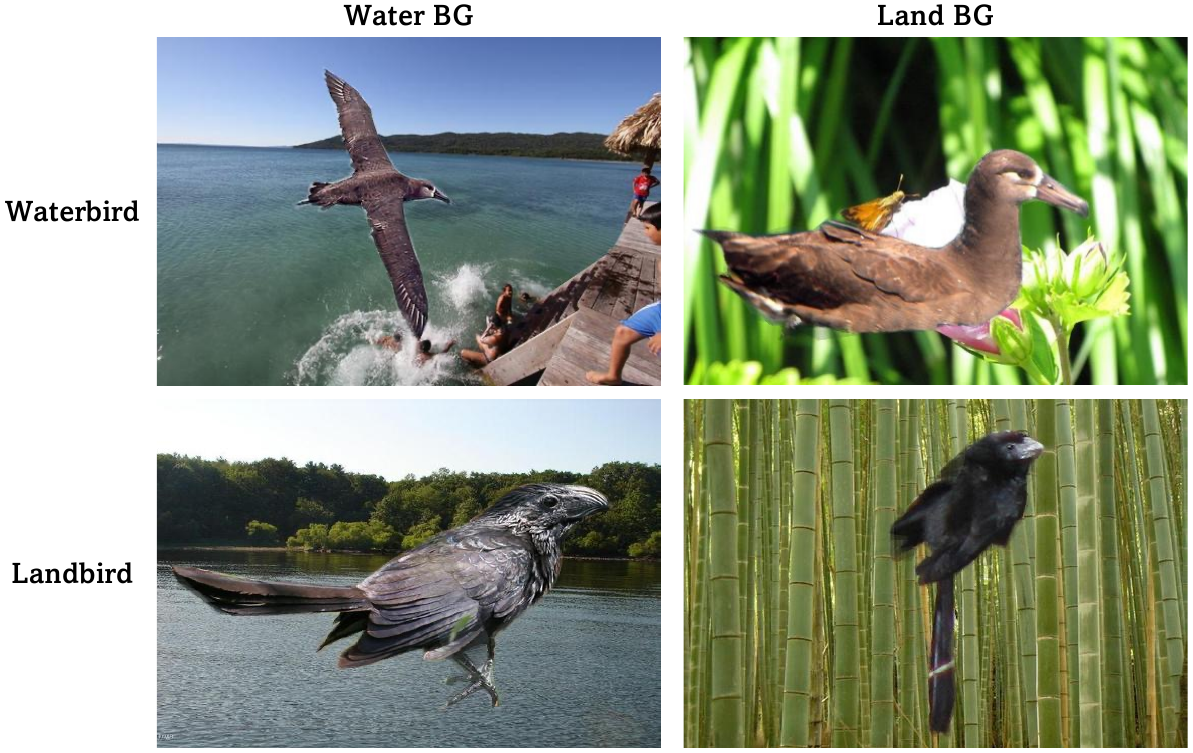}
\caption{\textbf{Sample images from Waterbirds.} 
}
\label{fig:sample waterbirds}
\end{figure}

\begin{figure}[t]
\centering
\includegraphics[width=0.87\linewidth]{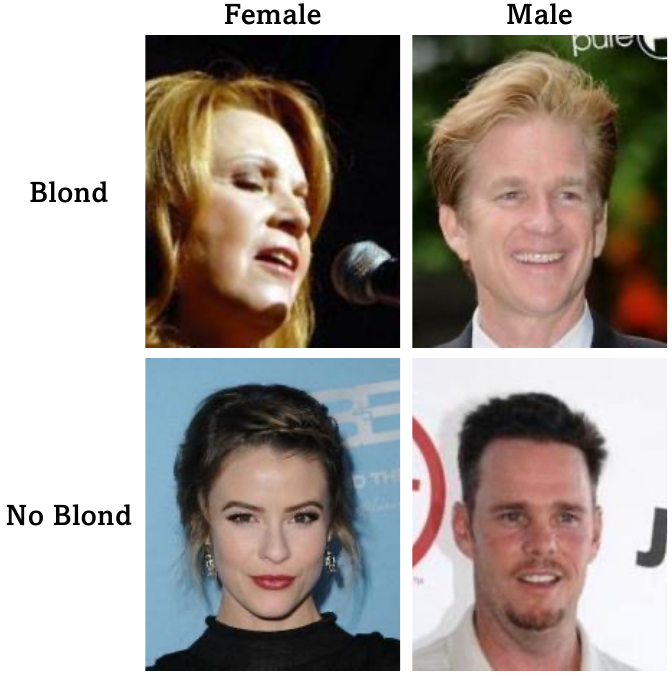}
\caption{\textbf{Sample images from CelebA.} 
}
\label{fig:sample celebA}
\end{figure}

\begin{figure}[t]
\centering
\includegraphics[width=0.97\linewidth]{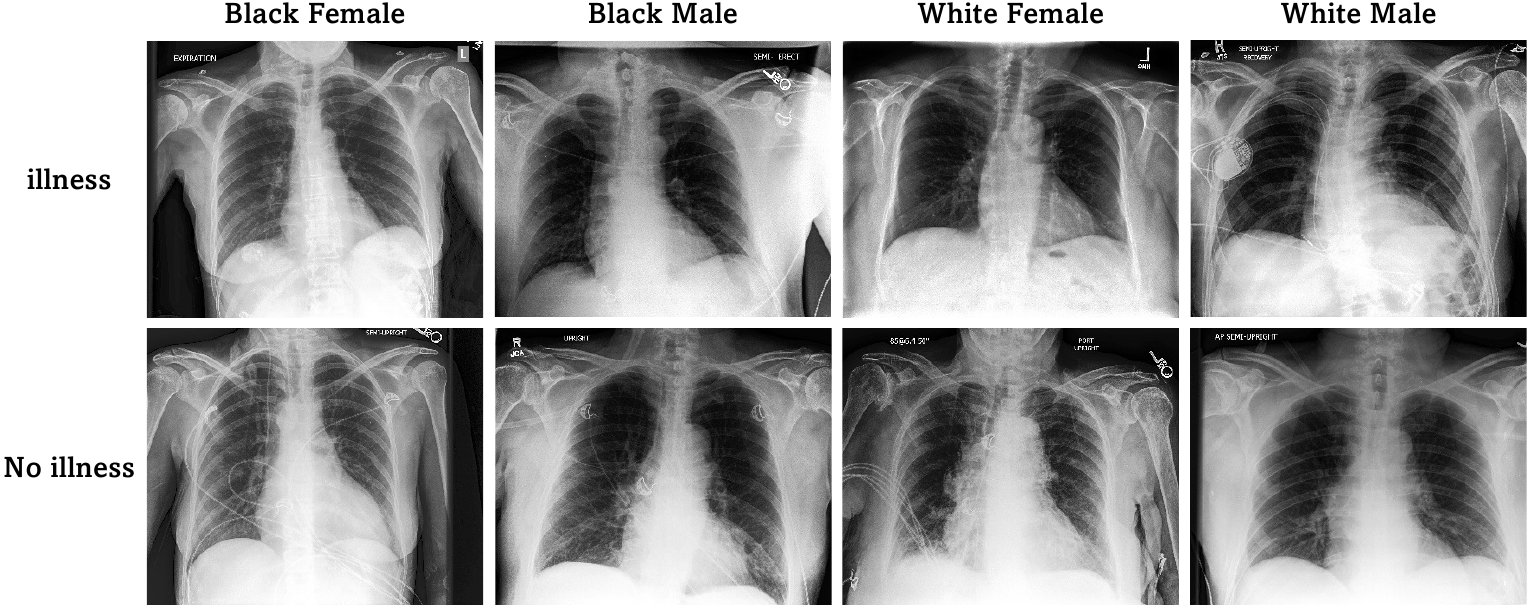}
\caption{\textbf{Sample images from CheXpert.} 
}
\label{fig:sample chexpert}
\end{figure}

\begin{figure}[t]
\centering
\includegraphics[width=0.87\linewidth]{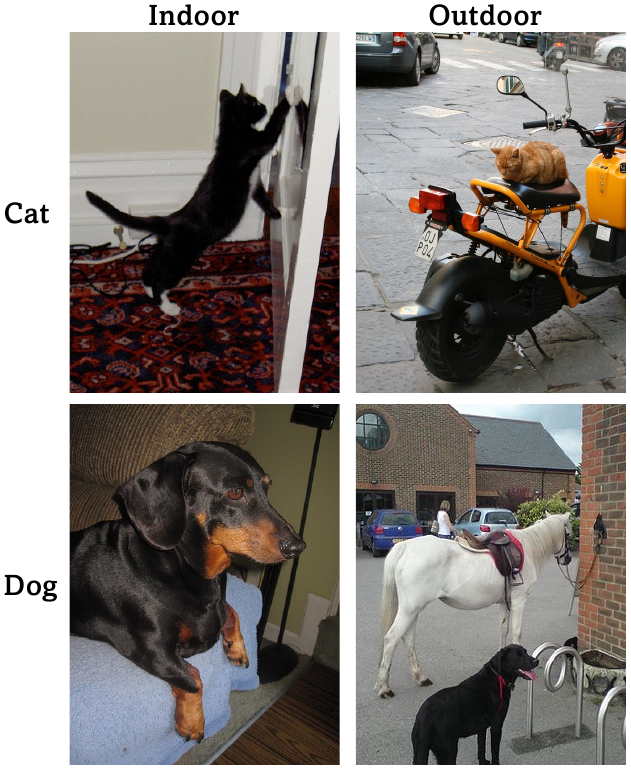}
\caption{\textbf{Sample images from MetaShift.} 
}
\label{fig:sample metashift}
\end{figure}

In this section we provide the details of the group information for all the datasets used in our experiments.
The groups and the number of samples in each group of Waterbirds, CelebA, CheXpert and MetaShift are summarized by Table~\ref{table:waterbirds group}, \ref{table:celebA group}, \ref{table:chexpert group} and \ref{table:metashift group}, respectively. 
Images samples from the datasets are shown in Figure~\ref{fig:sample waterbirds}, \ref{fig:sample celebA}, \ref{fig:sample chexpert} and \ref{fig:sample metashift}.

\end{document}